\documentclass[a4paper,fleqn]{cas-dc}

\usepackage[numbers]{natbib}

\usepackage{lineno,hyperref}
\usepackage{algorithm}
\usepackage{algorithmic}
\usepackage{amsmath}
\usepackage{booktabs}
\usepackage{float}
\usepackage{caption}
\usepackage{threeparttable}

\def\tsc#1{\csdef{#1}{\textsc{\lowercase{#1}}\xspace}}
\tsc{WGM}
\tsc{QE}
\tsc{EP}
\tsc{PMS}
\tsc{BEC}
\tsc{DE}

\begin{document}
\let\WriteBookmarks\relax
\def\floatpagepagefraction{1}
\def\textpagefraction{.001}

\shorttitle{Emotion detection for misinformation}

\shortauthors{Zhiwei Liu et~al.}

\title [mode = title]{Emotion Detection for Misinformation: A Review}                      



%

\author[1]{Zhiwei Liu}
\credit{Writing – original draft, Conceptualization, Methodology, Data curation, Visualization, review \& editing}

\author[1]{Tianlin Zhang}
\credit{Writing – original draft, review \& editing}
\author[1]{Kailai Yang}
\credit{Writing – original draft, review \& editing}
\author[1]{Paul Thompson}
\credit{Writing – original draft, review \& editing}
\author[1]{Zeping Yu}
\credit{Writing – review \& editing}

\author[1]{Sophia Ananiadou}
\credit{Writing – review \& editing}
\ead{sophia.ananiadou@manchester.ac.uk}
\cormark[1]

\ead[url]{https://research.manchester.ac.uk/en/persons/sophia.ananiadou}


\affiliation[1]{organization={National Centre for Text Mining, Department of Computer Science, The University of Manchester},
    city={Manchester},
    postcode={M1 7DN}, 
    country={UK}}

\cortext[cor1]{Corresponding author}
\cortext[cor2]{Principal corresponding author}

\begin{abstract}
With the advent of social media, an increasing number of netizens are sharing and reading posts and news online. However, the huge volumes of misinformation (e.g., fake news and rumors) that flood the internet can adversely affect people's lives, and have resulted in the emergence of rumor and fake news detection as a hot research topic. The emotions and sentiments of netizens, as expressed in social media posts and news, constitute important factors that can help to distinguish fake news from genuine news and to understand the spread of rumors. This article comprehensively reviews emotion-based methods for misinformation detection. We begin by explaining the strong links between emotions and misinformation. We subsequently provide a detailed analysis of a range of misinformation detection methods that employ a variety of emotion, sentiment and stance-based features, and describe their strengths and weaknesses. Finally, we discuss a number of ongoing challenges in emotion-based misinformation detection based on large language models and suggest future research directions, including data collection (multi-platform, multilingual), annotation, benchmark, multimodality, and interpretability.
\end{abstract}



\maketitle

\section{Introduction}

Misinformation is false information that is created specifically to mislead readers \cite{fernandez2018online1}, including \textit{fake news} and \textit{rumors}. Fake news refers to intentionally fabricated information whose publishing or dissemination may mislead readers or result in panic \cite{allcott2017social1}. Rumors are defined as unverified or unsupported hearsay or information that become spread among people \cite{bordia2007rumor2}. Rumors and fake news are now ubiquitous. They affect people's daily lives, alter their emotions and lead them to trust incorrect information. Social media platforms, such as Twitter, Facebook, Reddit and Sina Weibo, constitute important means not only for socialising, but also for spreading news and rumors, and generate a huge amount of information every day \cite{maniou2021semantic1}. According to the Datareportal April 2023 global overview\footnote{https://datareportal.com/reports/digital-2023-april-global-statshot}, approximately 4.80 billion people (about 60\% of the world's population) use social media. Moreover, its use is continuing to grow rapidly, with 150 million new user identities added in the last year, representing an annual growth rate of 3.2\%. Now that smartphones are commonplace, users can create, share and browse publicly available content on social media anytime and anywhere, thus increasing the ease and speed at which information can spread. However, due to a lack of effective regulatory measures, the Internet has become flooded with fake news and rumors, which can be challenging to distinguish from genuine facts \cite{scheufele2019science2}. Such misinformation can manipulate the emotions and intentions of netizens  \cite{li2021research3},  which in turn can impact upon social factors, politics and the economy. For example, during the COVID-19 pandemic, rumors about the virus spread across the Internet, which caused panic and tension among society \cite{cheng2021covid33}. Furthermore, recent advances in artificial intelligence (AI) and the emergence of large language models (LLMs) such as Instruct-GPT \cite{ouyang2022training7}, ChatGPT and GPT-4 \cite{openai2023gpt4} are making it increasingly straightforward to generate false information that appears highly convincing \cite{yan2023research4}. Accordingly, there is an urgent global-level need for methods that can detect misinformation effectively.

\begin{figure}[!t]
\centering
\includegraphics[width=\columnwidth]{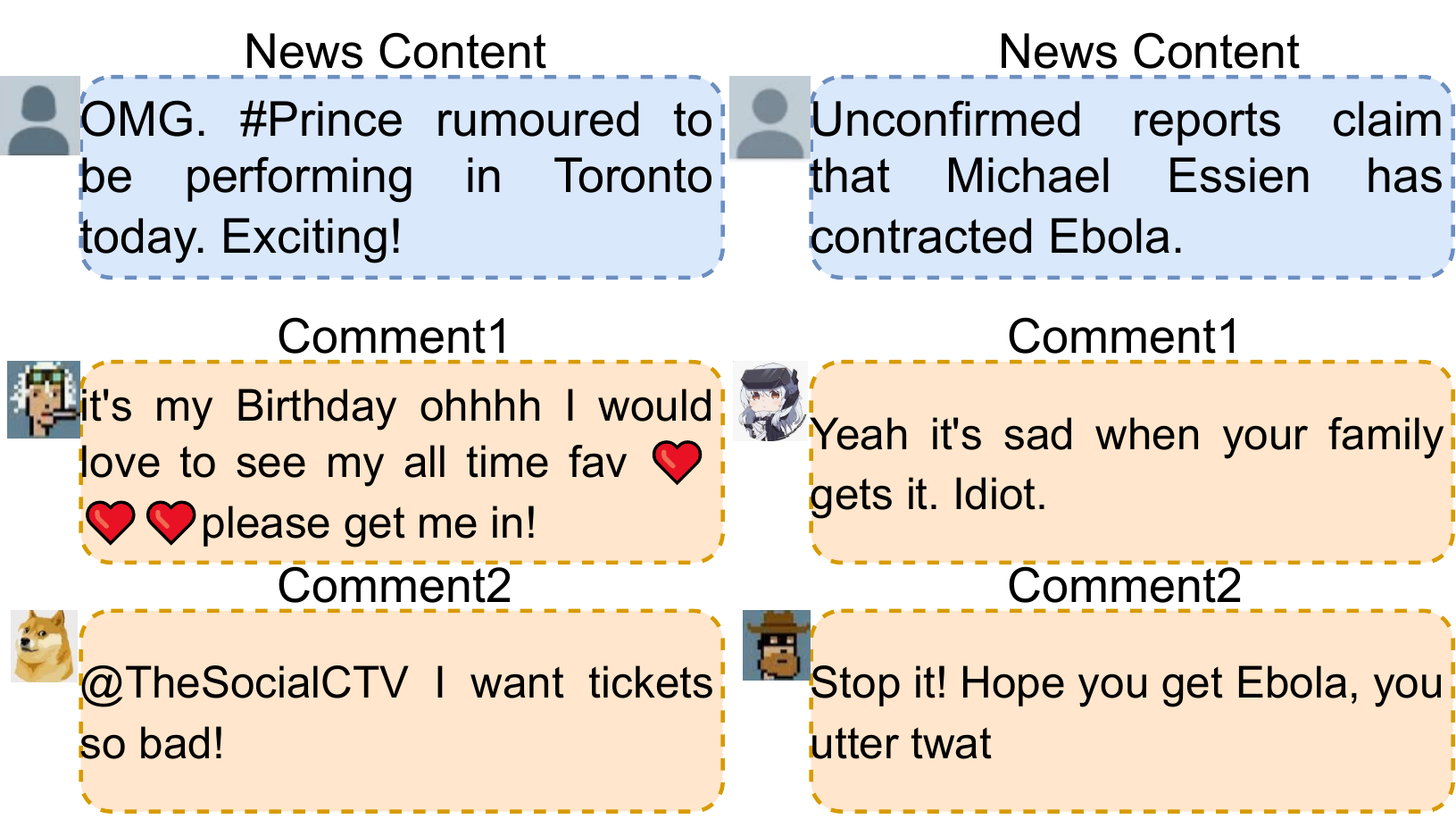}
\caption{Fake news samples}
\label{fakenewssamples}
\end{figure}

Rumors and fake news trigger specific emotions and sentiments. For example, Zaeem et al. \cite{zaeem2020sentiment333} observed a statistically significant relationship between negative sentiment and fake news. These emotions and sentiments can in turn give rise to specific behaviors or actions, such as the motivation to spread rumors  \cite{zhang2022emotional55}. Furthermore, readers are more likely to believe news that aligns with their existing beliefs \cite{horner2021emotions44}. For instance, in politics, conservative supporters are more likely to believe negative news about liberals. Rumor-mongering often takes advantage of these trends by disseminating fake news on social media channels that targets users with particular beliefs, and which triggers strong emotions \cite{li2021research3}. For example, fake news that attacks politics often intentionally embeds anger \cite{vosoughi2018spread}.   The aim of the rumor-mongers is to promote the further spread of the rumor by encouraging user actions such as forwarding, liking, and commenting. Such behaviour is exemplified in Figure~\ref{fakenewssamples}, which shows two samples of fake news on social media, with associated user comments. It has been found that false rumors tend to generate more reshares, spread over longer time periods, and become more viral when they include words that convey emotions of trust, anticipation or anger \cite{prollochs2021emotions213, prollochs2021emotions232}. Additionally, it was found that during the COVID-19 epidemic, there was a correlation between the level of anger felt by the public and the likelihood that rumors would be circulated  \cite{dong2020public123}. All of the above observations serve to demonstrate the strong relationships between emotions and misinformation. 

Recently, it has been shown that natural language processing (NLP) methods that recognise \textit {affective} information (e.g., emotions and sentiment \cite{cui2023surveysentiment}) in text can make important contributions towards the automated detection of misinformation and conspiracies \cite{alonso2021sentiment12}.
Significant advances in many NLP tasks (e.g., classification, summarisation, question answering, and information extraction) have been facilitated by the advent of deep learning (DL) methods, which are able to extract higher-level and more complex feature representations through multiple processing layers, compared to conventional machine learning methods. Various DL approaches that exploit emotion features have been used to approach the problem of misinformation detection, including Convolutional Neural Networks (CNN) \cite{luvembe2023dual14}, Recurrent Neural Networks (RNN) \cite{iwendi2022covid11} and Graph Convolutional Networks (GCN) \cite{wang2021rumor151}. Furthermore, pre-trained language models such as BERT\cite{miao2021syntax21}, RoBERTa \cite{kolev2022foreal321}, and LLMs \cite{hu2023bad1,pavlyshenko2023analysis2,cheung2023factllama3} have been used as backbone models for detecting misinformation. The various proposed methods have exploited emotion features in diverse ways. For example, Al-Saif et al \cite{al2023exploring} developed a context-aware approach for rumor detection in Arabic social media that combines emotion features with other types of features (i.e., topics and reactions), while Zhang et al. \cite{zhang2021miningwww} accounts for the \textit{dual emotions} expressed in both the fake news post and its follow-up comments. Emotion detection can also be successfully employed as an auxiliary task within a multi-task framework to improve the accuracy of fake news detection \cite{choudhry2022emotion}. Such examples illustrate the potential for emotion information to be integrated within misinformation detection methods in a broad range of ways to improve performance.

Determining the  \textit{stance} of social media users towards news also plays a crucial role in identifying misinformation \cite{yaakub2019review111, santhoshkumar2020earlier2, tian2020early3}. Stance is defined as the expression of an attitude towards a given piece of information \cite{conforti2018towards}, which may include supporting, denying, querying, or commenting upon it \cite{derczynski2017semeval}. Users often take a stance towards rumors propagated in online spheres \cite{zojaji2022adaptive4}. For example, the public has expressed various attitudes towards climate change on social media platforms \cite{upadhyaya2023towards}.  Moreover, users are more likely to accept and support information that aligns with their viewpoints \cite{kim2019combating}. For instance, individuals with strong opinions about "Americanness" tended to demonstrate support in their tweets relating to former US President Trump's October 2018 post concerning the cancellation of birthright citizenship \cite{worrall2022sentiment}. Emotions and sentiment have an underlying connection with attitudes and are thus advantageous for stance detection \cite{mohammad2017stance6,sobhani2016detecting2}. For example, if a person expresses positive feelings towards a political candidate, then this is likely to indicate that they support or agree with the candidate's policies. The importance of sentiment and emotion has been confirmed by a number of studies that have used them in combination with other features for stance detection in rumors and fake news\cite{hanselowski2018retrospective88,ghanem2019upv11,khandelwal2021fine222}. Examples include Wang et al. \cite{wang2017ecnu55}, who combined emotion and sentiment with Twitter metadata features, Xuan et al. \cite{xuan2019rumor111}, who integrated emotion with content and user features, and Parimi et al. \cite{parimi2023flacorm11}, who made use various features of rumors, including content and emotions.  

Several surveys relating to rumor and fake news detection have been published recently. The majority of these reviews a variety of detection techniques \cite{liu2023review1, 10176530review2, 10182240review3, choudhary2021review4, bondielli2019survey5}, while \cite{varlamis2022survey6} focuses specifically on methods that employ GCNs. D’Ulizia et al. \cite{d2021fake7} provides an overview of available datasets for evaluating fake news detection methods. Alsaif et al. \cite{alsaif2023review8} reviews recent approaches that use stance detection as a means to identify rumors, while Hardalov et al. \cite{hardalov2021survey9} examines the relationship between stance detection and misinformation identification. Shahid et al, \cite{shahid2022you10} conducts a comprehensive survey of state-of-the-art methods for detecting malicious users and bots. Shelke et al. \cite{shelke2019source11} analyzes methods for detecting sources of misinformation in social networks. Among these studies, little attention is paid to the role of emotion in fake news and rumor detection. Although Alonso et al. \cite{alonso2021sentiment12} provide an overview of the application of sentiment analysis in the detection of fake news, it touches only very briefly on approaches that employ fine-grained emotion information, and does not discuss emotion-based stance detection. 
Furthermore, given the highly active nature of research in this area, there are many recently developed methods that are not included in the above review. To the best of our knowledge, the current article constitutes the first comprehensive survey of methods that use both sentiment and emotion as a means to detect fake news, rumors, and stances. The aim of the survey is to facilitate an enhanced understanding about the latest developments in this area and to act as a driver and a guide for promising future research.

We collected articles from five different literature search platforms, i.e., IEEE Xplore, ACM Digital Library, Web of Science, Scopus, and DBLP. The process for article selection process consisted of three main steps, i.e.: \textit{collection, preliminary screening}, and \textit{manual review}. 

\textbf{Collection:} Similarly to the search strategy described in \cite{zhang2022natural,zhang2023emotion1}, we conducted an initial keyword search aimed at retrieving articles published between January 2016 and September 2023 that mention both misinformation and affective information. The specific query used was as follows: \textit{(emotion OR sentiment OR affective) AND (rumor OR "fake news" OR misinformation OR disinformation)}. The search resulted in the retrieval of 6,483 articles.  

\textbf{Preliminary screening:} After deduplication, we employed RobotAnalyst \cite{przybyla2018prioritising}, a tool that prioritizes articles based on relevance feedback and active learning \cite{o2015using2,miwa2014reducing3} in order to minimize the amount of human work required in the screening phase of reviews. Articles were screened based on title and abstract, and were retained only if: (1) They were relevant to rumor/fake news analysis or detection. (2) They involved the use of affective information. The screening process resulted in the identification of 473 articles for further review. 

\textbf{Manual review:} We conducted a manual full-text examination of the articles resulting from the preliminary screening phase, and retained those that: (1) Focus on methods both for analyzing or detecting misinformation, \textit{and} detecting emotions and/or sentiment. (2) Apply learning methods to the task of misinformation detection or analysis. (3) Use affective information as a feature for misinformation detection or analysis. By applying these criteria, 90 articles were retained, and form the basis for the detailed analysis presented in this review. 

\begin{figure}[!t]
\centering
\includegraphics[width=\columnwidth]{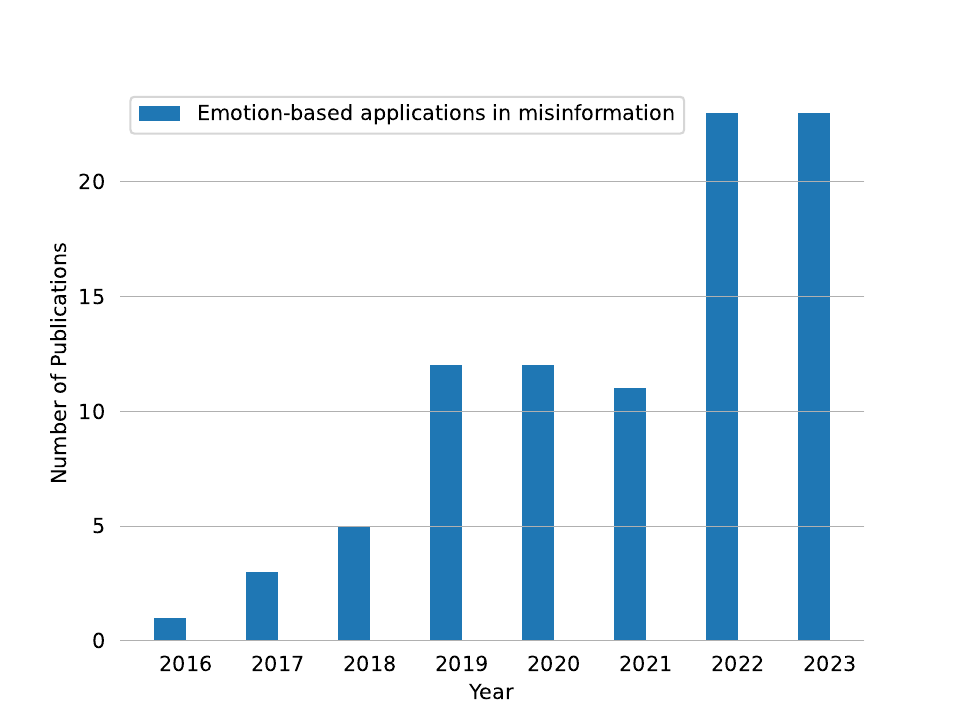}
\caption{Distribution of publications on emotion-based applications in misinformation since 2016.}
\label{Literature_statistics}
\end{figure}

Figure~\ref{Literature_statistics} illustrates the temporal distribution of studies describing emotion-based applications in misinformation that have been published in recent years. Particularly noticeable is the significant surge in the number of articles published over the last two years. This provides evidence of an increasing appreciation of the importance of emotion in detecting rumors and fake news. 

In this review, we focus on advanced emotion-based fusion methods for misinformation detection. Our main contributions are as follows:

1. We summarize the findings of articles exploring relationships between emotions and misinformation, in order to motivate emotion-based approaches to misinformation detection.

2. We categorize and summarize available datasets that can support misinformation detection.

3. We categorise and discuss emotion-based methods for misinformation detection based on both conventional machine learning and DL methods, with a focus on advanced emotion-based fusion approaches. We also provide an overview of articles concerning emotion-based stance detection in misinformation.

4. We present and analyze the performance of the advanced methods discussed, and discuss their relative strengths and weaknesses.

5. We outline a number of challenges faced in the development of misinformation detection methods, and suggest promising future research directions, with an emphasis on the increasingly important role of  LLMs.

\begin{figure*}[!t]
\centering
\includegraphics[width=1.5\columnwidth]{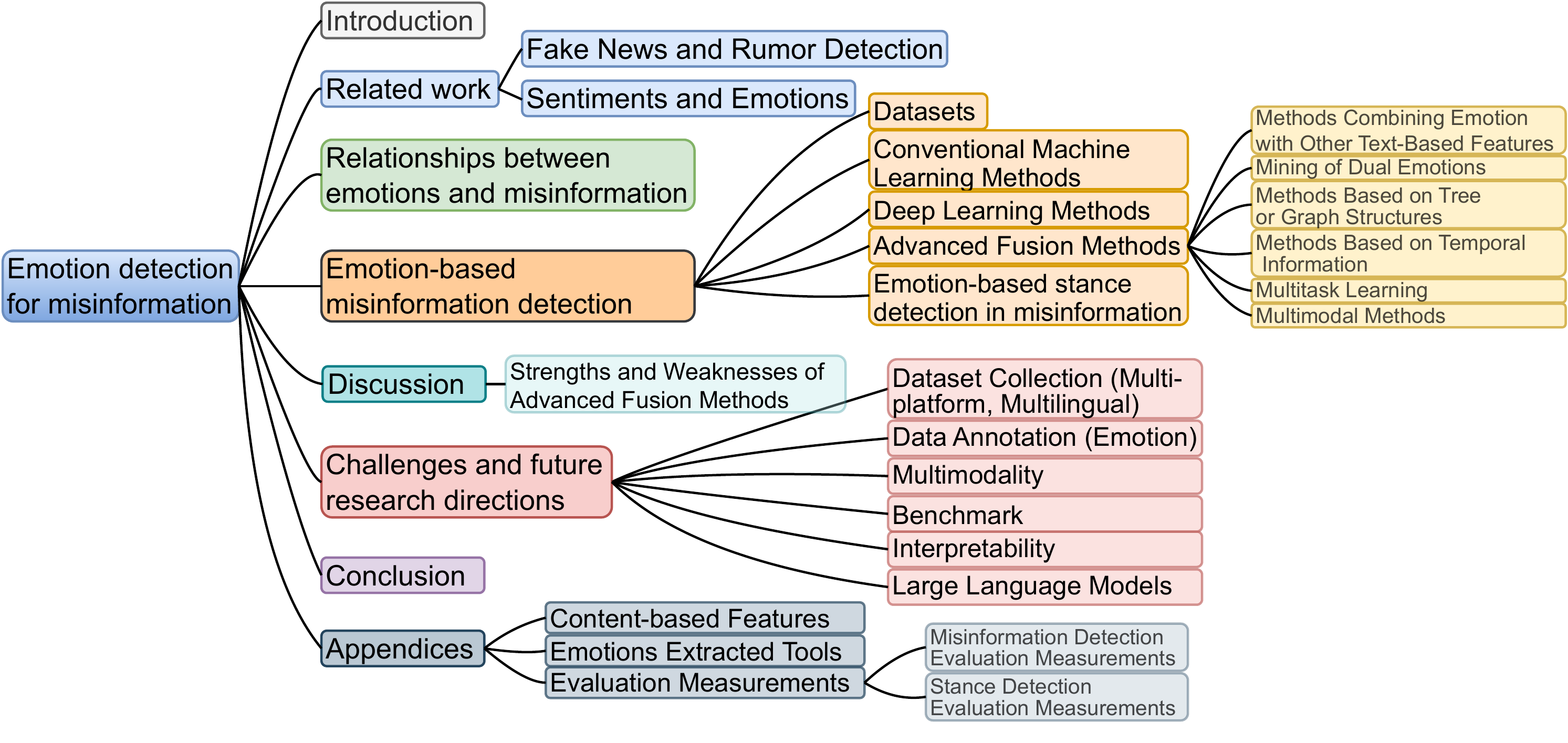}
\caption{Structure of this article}
\label{diagramthispaper}
\end{figure*}

Figure~\ref{diagramthispaper} illustrates the structure of the remainder of the article, which may be summarised as follows: Section \ref{sec:relatedwork} introduces related work on rumors, fake news detection, and emotion detection. Section \ref{sec:relationship} presents studies that analyze relationships between emotions and misinformation. Section \ref{sec:emotion-basedrumordetec} explores approaches to emotion-based misinformation detection, including a summary of available datasets, a detailed analysis of advanced fusion methods and a summary of emotion-based stance detection in misinformation. Section \ref{sec:disscussion} discusses the strengths and weaknesses of advanced emotion-based misinformation detection methods. Section \ref{sec:challengeanddirection} presents ongoing challenges and future research directions; Finally, Section \ref{sec:conclusion} concludes this paper by summarizing our findings.

\section{Related work \label{sec:relatedwork}}

\subsection{Fake News and Rumors Detection}

The convenience of accessing social media platforms on various electronic devices means that people can easily post or access large amounts of information on the Internet. This can lead to the rampant spread of misinformation. Certain individuals intentionally spread rumors to gain attention, mislead readers, or make a profit, even though such rumors can pose significant harm to society \cite{wu2016mining5}. Therefore, there is an urgent need to detect misinformation in an efficient and effective manner. A large body of research has aimed to respond to this need, which has been summarized in various reviews, which cover both methods for detecting rumors and fake news \cite{liu2023review1, 10176530review2, 10182240review3, choudhary2021review4, varlamis2022survey6} and potential applications of these methods, including source detection \cite{shahid2022you10,shelke2019source11}, bot detection \cite{shahid2022you10,long2022method2, chawla2023hybrid3} and stance detection \cite{alsaif2023review8,hardalov2021survey9}.

\begin{figure*}[!t]
\centering
\includegraphics[width=1.5\columnwidth]{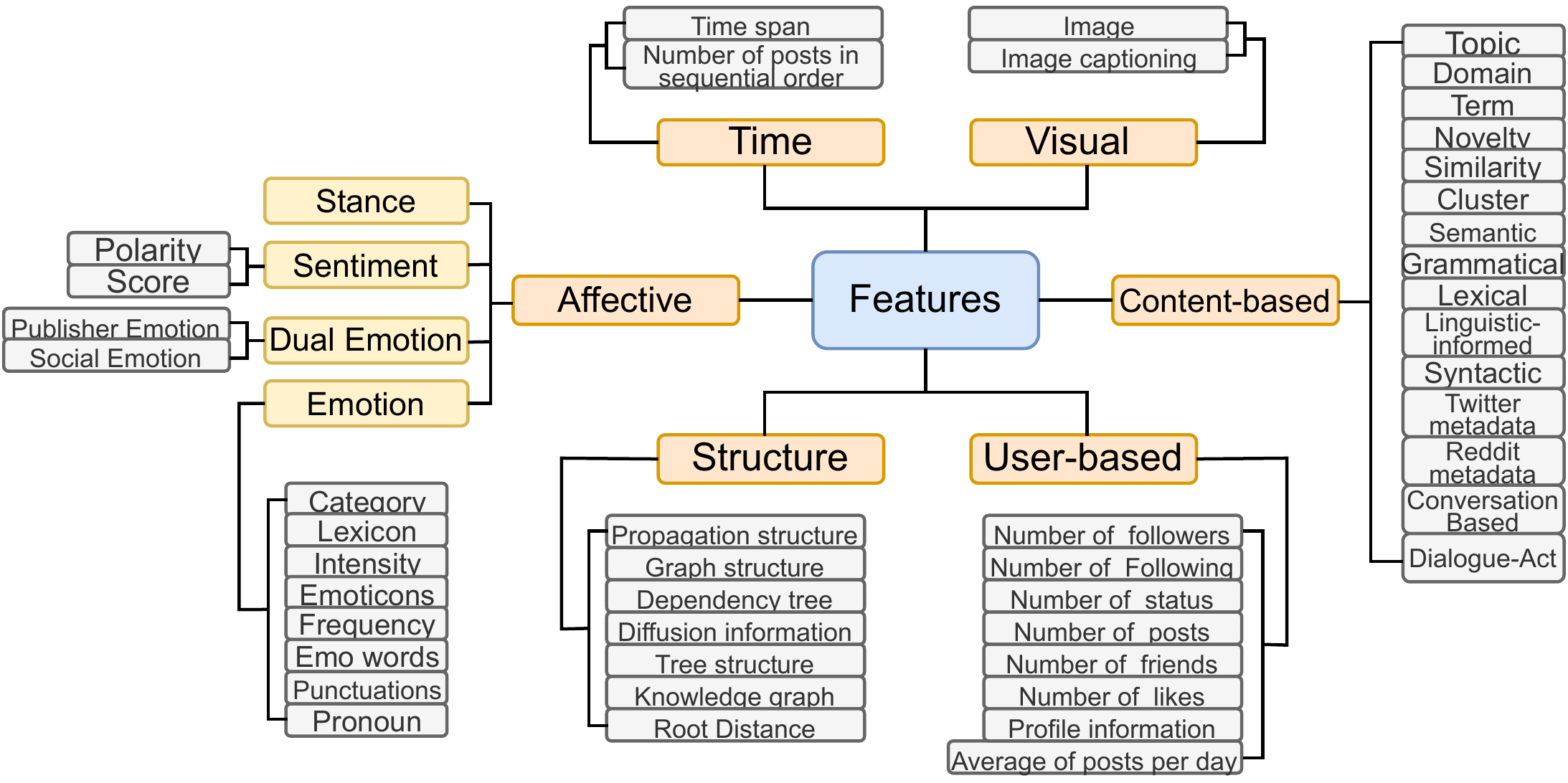}
\caption{The features used in rumor and fake news detection}
\label{detectionfeatures}
\end{figure*}

Misinformation detection approaches consist of three main components, i.e., the \textit{datasets} used to support their development, the \textit{methods} used to perform detection, and the \textit{features} used within these methods. The majority of the datasets are obtained from social media platforms such as Twitter, Facebook and Sina Weibo, or from fact-checking websites, such as Snopes\footnote{https://www.snopes.com/}, Factcheck\footnote{https://www.factcheck.org/}, PolitiFact\footnote{https://www.politifact.com/}. Detection methods may be divided into those based on conventional machine learning \cite{choudhary2021review4} or DL \cite{10182240review3,varlamis2022survey6}. Figure~\ref{detectionfeatures} presents a range of features that have been employed for misinformation detection. Among these features, content-based features constitute the most diverse class; Table~\ref{tab:contentbasedfeatures} in Appendix \ref{appendix:features} provides more detail regarding the specific types of features that fall under each of the content-based sub-classes shown in Figure~\ref{detectionfeatures}. Within the \textit{Affective} group of features, \textit{dual emotion} features aim to account for the importance of considering different emotional perspectives when identifying misinformation, i.e., both  \textit{publisher emotion}, which refers to the emotions conveyed in an original post that starts a thread on social media, and \textit{social emotion}, which refers to the emotions expressed in follow-up posts that respond to and/or comment on the original post.

\subsection{Sentiments and Emotions}

Sentiments and emotions are important and fundamental aspects of our lives. What we do and say reflects our emotions in some way. Emotion detection (ED) and sentiment analysis (SA) are two types of NLP techniques for analyzing human expressions that can help us to understand people's feelings towards specific topics \cite{8004002Hakak2017}. SA \cite{devika2016sentiment2} aims to capture the overall emotional tone conveyed by a data source (usually \textit{positive}, \textit{negative}, or \textit{neutral}), along with the strength of this tone \cite{thelwall2010sentiment}. ED is the process of classifying data at a finer-grained level, according to the emotions that it conveys. Compared to sentiment, the term \textit{emotion} refers to more specific and stronger feelings \cite{uymaz2022vector3}. For example, positive sentiment encompasses a range of different emotions, such as  \textit{happiness} and \textit{joy}, while negative sentiment includes the emotions of \textit{sadness} and \textit{anger}, among others. 

A number of theoretical emotion classification models has been proposed, which can be divided into two categories, i.e., \textit{categorical} and \textit{dimensional} \cite{sailunaz2018emotion3}. Categorical models define a single discrete set of emotional states; examples include Shaver \cite{shaver1987emotion1} (sadness, love, joy, anger, surprise and fear), Ekman \cite{ekman1992argument2} (joy, anger, fear, disgust, sadness and surprise), and Plutchik \cite{plutchik1980general3} (anticipation, surprise, anger, fear, trust, disgust, joy and sadness). In contrast, dimensional models posit that emotions can be decomposed into a number of distinct dimensions.  One of the best-known examples is Plutchick’s \textit{wheel of emotions} \cite{plutchik1980general3,acheampong2020text4}, in which emotions are defined in a two-dimensional space of \textit{valence} and \textit{arousal}. The wheel is divided into 24 primary, secondary, and tertiary dyads based on eight basic emotions. Other popular dimensional emotion models include the PAD model \cite{russell1977evidencepad}, which is based on three dimensions, i.e., \textit{Pleasure} (the pleasantness of the emotion), \textit{Arousal} (the level of physiological activation or intensity of the emotion), and \textit{Domination} (the degree of control or dominance experienced in the emotion); and the VAD model \cite{warriner2013normsvad}, in which \textit{Arousal} and \textit{Dominance} are supplemented by \textit{Valence} (the positivity or negativity of the emotion). 

A range of automated methods has been developed to detect both sentiment and emotions in text, which may be broadly categorized into dictionary-based, conventional machine learning \cite{alslaity2022machine1}, and DL \cite{peng2022survey2} methods. The dictionary-based approach involves constructing an inventory of words that denote specific sentiments and/or emotions, and matching them against words appearing in the text to be processed to obtain information about the sentiments and emotions conveyed. Meanwhile, methods based on conventional machine learning and DL apply learning algorithms to datasets annotated with sentiment or emotion labels to teach them how to detect the different ways in which these types of emotions may be conveyed in text. Recently, there has also been a growing interest in exploring how LLMs can be exploited to enhance the accuracy of sentiment analysis and emotion detection \cite{zhang2023enhancing,feng2023affect2,lei2023instructerc1}.

\section{Relationships between emotions and misinformation \label{sec:relationship}}

Although emotions are regarded as a dominant driver of human behavior, the exploration of their role in the online diffusion of misinformation has only recently begun. Misinformation can evoke emotional responses in readers, which in turn can lead to specific behaviors, such as belief in the information, resharing or liking it, etc \cite{horner2021emotions44}. 

Table \ref{tab:relationship} lists a range of recent studies that has investigated the relationships between emotions and misinformation, e.g., how the expression of particular emotions can indicate that a data source is likely to contain misinformation and/or predict the likely response of readers. For each study, we list the dataset used, the ED and relationship analysis methods (RAM) methods employed, and details of the most important relationships identified.  The most commonly explored topic is COVID-19, according to the explosion of rumors and fake news generated by the pandemic. To perform ED, the majority of researchers applied dictionary-based methods (Table~\ref{EmotionsExtractedTools} for details) or traditional machine learning methods, while Wu et al. \cite{wu2022emotion20333} manually annotated discrete emotions based on the Pleasure-Arousal-Dominance (PAD) emotional state model. In \cite{horner2021emotions44,zhang2022emotional55,martel2020reliance22,rijo2023s11,tan2023application33,ali2022effects667}, questionnaires were designed to ask participants to directly report their emotions. Among these approaches, Zhang et al. \cite{zhang2022emotional55} and Martel et al. \cite{martel2020reliance22} use the Positive-Negative Emotional Scale (PANAS) to further quantify the emotional state of participants. The analysis of Li et al. \cite{li2023multi2223} was based on the results obtained from their novel Multi-EmoBERT multi-label emotion recognition tool. Wan et al. \cite{wan2023fake222} used a mixture of existing NLP tools and lexicons for ED, enhanced using rules and automated weighting. They extracted Emotion Triple Elements to study potentially different responses to emotional triggers.  For relationship analysis, a range of commonly used statistical analysis methods has been applied, including Logistic Regression (LR) \cite{wu2022emotion20333,prabhala2019emotions444}, Linear Regression \cite{chuai2022anger33}, and T-Test \cite{zhou2023does22,zhou2023does222}.

Various indicators have been used to judge the impact of emotions on the spread of rumors, the degree of outbreak, etc. For example, the questionnaires of \cite{zhang2022emotional55,rijo2023s11,ali2022effects667} directly asked participants what actions they would take when faced with certain types of news, such as sharing intention or ``likes''. Other studies used cascade size, cascade lifetime, and structural virality \cite{prollochs2021emotions213,prollochs2021emotions232,solovev2022moral555} to analyze the patterns of misinformation spread. Cascade size corresponds to the number of forwardings generated by a cascade; cascade lifetime is the length of time that a rumor cascade remains active, i.e., the time elapsed between the root broadcast and the final forwarding; structural virality \cite{goel2016structural} provides an aggregated metric combining the depth and breadth of a cascade. In addition, many studies have analyzed relationships by investigating the number of rumors that occur over time, or by comparing the number of rumors that convey different emotions.

The analyses detailed in Table \ref{tab:relationship}  reveal a number of important relationships between emotions and misinformation, which can sometimes depend on the types of topics being discussed. Misinformation is generally associated with a significant level of high-arousal emotions such as anger, sadness, anxiety, surprise, and fear. Rumors conveying anger, sadness, anxiety, and fear are likely to generate a large number of shares, and to be long-lived and viral \cite{prollochs2021emotions213,dong2020public123,wu2022emotion20333,prabhala2019emotions444}, while emotional appeals (like anger and disgust) can increase users’ engagement with fake posts \cite{sui2023falsehood333}. Fake news expresses higher overall emotion, negative sentiment, and lower positive sentiment than genuine news \cite{zhou2023does22,zhou2023does222}. In general, it may be concluded that sentiment, emotions, and misinformation are inextricably intertwined, thus confirming that sentiment and emotion both have important parts to play in the automated detection of fake news and rumors.

\begin{table*}
\footnotesize
\caption{Relationships between emotions and misinformation. ED: Emotion Detection, RAM: Relation analysis methods. MANOVA: Multivariate Analysis of Variance,
MANCOVA: Multivariate Analysis of Covariance, ANOVA: Analysis of Variance.}
\label{tab:relationship} 
\begin{tabular}{p{0.2cm}p{0.3cm}p{1.6cm}p{2.5cm}p{2.7cm}p{8cm}}
\toprule
Pub                      & Year & Data                   & ED                                    & RAM                                          & Relationship (Partly)                                                                                                                     \\
\midrule

\cite{prabhala2019emotions444}   & 2019 & Demonetization related & LIWC                                    & Logistic Regression                                         & Posts with a higher level of anger, sadness, and anxiety are indicative of rumor.                                                       \\
\cite{dong2020public123}         & 2020 & COVID-19 Related       & Manual                                  & Time-lagged Cross-correlation Analyses                      & The angrier, sadder, or more fear the public felt, the more rumors there were likely to be.                                                  \\
\cite{martel2020reliance22}      & 2020 & News Headlines              & Questionnaire,PANAS                    & Linear Mixed-effects Analyses                               &  Emotion plays a causal role in people’s susceptibility to incorrectly perceiving fake news as accurate.                                                                                        \\
\cite{fersini2020profiling111}   & 2020 & \cite{clef2020-checktahtlab}              & NRC                                     & SVM                                                         & Emotion-based features contribute more to the rumor recognition capabilities than personality-based ones.                           \\
\cite{zaeem2020sentiment333}     & 2020 & Open-Source Data       & Meaningt.Ioud, TextBlob5, AFINN6        & Chi-square Test, P(T|S), Goodman and Kruskal's Gamma        & Relationships exist between negative sentiment and fake news, and between positive sentiment and genuine news.                         \\
\cite{prollochs2021emotions213}  & 2021 & Twitter                & Questionnaire                           & Generalized Linear Model                                    & Rumors conveying anticipation, anger, or trust, or which are highly offensive, generate more shares, are longer-lived, and more viral.            \\
\cite{prollochs2021emotions232}  & 2021 & Twitter                & NRC                                     & Generalized Linear Model                                    & False rumors with a high proportion of terms conveying positive sentiment, trust, anticipation, or anger are more likely to go viral.     \\
\cite{wang2021sentiment05}       & 2021 & COVID-19 Related       & Decision Tree                           & SPSS 22.0, Granger Causality Test                           & The more negative people feel about COVID-19, the more likely it is that rumors will be generated.                                                      \\
\cite{horner2021emotions44}      & 2021 & News Headlines              & Questionnaire                           & MANOVA, MANCOVA, ANOVA                                      & Emotional reactivity of participants is associated with response behavior intentions.                                                    \\
\cite{zhang2022emotional55}      & 2022 & Questionnaire          & Questionnaire, PANAS                    & Multilevel Linear Regression                                & Expression of emotion in online rumors positively affects readers' emotions. Readers’ emotions affect their intentions to spread rumors. \\
\cite{wu2022emotion20333}        & 2022 & COVID-19 Related       & Pleasure-Arousal-Dominance                                     & Logistic Regression                                         & Weibo messages filled with high-arousal emotions such as fear,   anger and surprise are more likely to be rumors.                \\
\cite{chuai2022anger33}          & 2022 & Twitter, Weibo         & Emotion Lexicon, ML, DL                 & Logit Regression, Linear Regression                         & The ease with which fake news is spread online is positively associated with the strength of anger that it conveys.                                     \\
\cite{khan2022exploration222}    & 2022 & Open-Source Data       & Textblob                                & Histogram                                                   & The negativity score of fake news is slightly higher than that of real news.                                                            \\
\cite{solovev2022moral555}       & 2022 & COVID-19 Related       & NRC                                     & Generalized Linear Model                                    & False rumors that include a large number of emotion words condemning others are more viral.                                                  \\
\cite{ali2022effects667}         & 2022 & Anti-vaccination       & Questionnaire, LIWC                     & ANCOVA, SPSS24                                              & Individuals who are more neutral towards vaccines and are angry are more likely to believe and share anti-vaccine fake news compared with individuals who have anti-vaccine attitudes and are fearful.         \\
\cite{rijo2023s11}               & 2023 & Political              & Questionnaire                           & SPSS29, Structural Equation Modelling                       & Negative beliefs about the political system increases emotional response to both genuine and fake news.                                        \\
\cite{zhou2023does22}            & 2023 & COVID-19 Related       & TextBlob,cn-sentiment -measures, LIWC   & T-Test                                                      & Fake news expresses a higher level of overall emotion,   negative emotion, and anger than real news.                                           \\
\cite{tan2023application33}      & 2023 & COVID-19 Related       & Questionnaire                           & Partial Least Squares, Multigroup Analysis                  & Surprise is felt most intensely towards celebrity fake news and the toilet paper shortage rumor.                               \\
\cite{li2023multi2223}           & 2023 & Open-Source Data       & Multi-EmoBERT Fig.~\ref{detectwithtree} & Chi-square Test                                             & Fake news is  associated with negative emotions and co-existing emotions in certain contexts.                                            \\
\cite{wan2023fake222}            & 2023 & COVID-19 Related       & Emotion Triple Elements                 & Inter-rater Reliability Analysis, Cohen’s Kappa Coefficient & The emotion of fear plays an important role in the spread of fake news.                                                                        \\
\cite{gagiano2023emotionally111} & 2023 & SouthAfrican Website  & VADER, T5 transformer                   & Word Clouds, T-SNE Plots and Histograms                     & Fake news in South Africa conveys more anger, joy, sadness and fear than genuine news.                                                     \\
\cite{zhou2023does222}           & 2023 & COVID-19 Related       & TextBlob, cn-sentiment-measures         & T-Test                                                      & Fake news contains higher overall emotion, negative emotion,   and anger than real news.                                                  \\
\cite{sui2023falsehood333}       & 2023 & COVID-19 Related       & IBM Watson’s NLU                        & Chi-square Test, Linear Regression                          & Anger and disgust increase users’ engagement with fake posts.    \\                                                                      \bottomrule                  
\end{tabular}
\end{table*}

\section{Emotion-based misinformation detection \label{sec:emotion-basedrumordetec}}

Motivated by the results of analyses such as those outlined in Section \ref{sec:relationship}, many studies have used sentiment and/or emotions as the main features to guide the automated detection of fake information. In this section, we provide a detailed survey of emotion-based methods for misinformation. We firstly introduce the datasets used to support the development of such methods, and subsequently describe a range of detection methods employing a mixture of conventional machine learning, DL methods, and advanced fusion techniques. We additionally provide a summary of the closely related task of emotion-based stance detection in misinformation. Table \ref{tab:methods} lists the complete set of emotion-based misinformation detection methods that we have reviewed. Appendices \ref{appendix:EmotionsExtractedTools} and \ref{appendix:EvaluationMeasurements}, respectively, list commonly used ED tools and provide an overview of evaluation metrics used in misinformation detection.

\subsection{Datasets}
Table \ref{tab:datasum} lists a range of publicly available datasets aimed at facilitating the development and/or evaluation of misinformation detection methods. The majority of these datasets consist of data obtained from popular social media platforms and fact-checking websites, such as Twitter, Weibo, Reddit, politifact.com, gossipcop.com, etc. For each dataset, we provide its commonly used name and reference, its source, a description of its size and main characteristics, its level of availability, and notes. The latter is used to indicate datasets that cover languages other than English, those that are multimodal, those specifically concerning COVID-19, and those annotated with stance information. Datasets without notes consist of textual English data that is labeled according to whether or not it represents misinformation.   

As may be observed in Table \ref{tab:datasum}, the majority of datasets is publicly available, which is highly advantageous to promote research in the field of misinformation. Due to the prevalence of rumors relating to the COVID-19 pandemic on social media, there has been a trend towards collecting misinformation datasets relating to this topic, as a means to explore rumor detection in the field of health disease transmission \cite{patwa2021fightinginfodemic,cui2020coaid,COVID-19infodemic2020,arora2022modified232}.
Another important feature of several of the datasets listed (including  PHEME\cite{Kochkina2018pheme}, the Twitter series \cite{ma2017detectnewtwetter1516}, and the Weibo series \cite{zhang2021miningwww,ma2016detectingtwitter16,nan2021mdfendweibo21}) is that they include comments/replies relating to original news stories or tweets. Such datasets allow the exploration of methods that take into account the dual publisher and social emotions, and the possible interactions between them, to improve the accuracy of misinformation detection. We can furthermore observe that several datasets are multimodal, i.e., they consist of both text and images. These include FakeNewsNet\cite{shu2020fakenewsnet}, Fakeddit \cite{nakamura2020fakeddit} and MediaEval2016 \cite{inproceedingsMediaEval2016}. The inclusion of images in these datasets provides scope to explore methods that take advantage of visual clues to complement text-based information in identifying misinformation. Although the majority of datasets only contain English text, there is a growing number of corpora that cover other single and sometimes multiple languages, including Chinese, Portuguese, Spanish and Danish, thus providing opportunities to develop methods that are multilingual and/or which target lesser resourced languages. While most datasets are annotated according to whether or not their constituent text constitutes fake news or rumor, there is also a number of corpora annotated with stance-related labels, which can facilitate investigations into how stance information can contribute towards the detection of misinformation.

\begin{table*}[htb]
\footnotesize
\caption{Summary of misinformation datasets. A: Available, N: No link, R: Request. An empty cell in the \textit{Notes} column means that the dataset is English and consists only of textual data}
\label{tab:datasum} 
\begin{tabular}{p{2.3cm}p{0.9cm}p{10cm}p{0.5cm}p{2cm}}
\toprule
Dataset                                                                             & Source          & Description                                                                                                                                                                                                                                                                                                                       & A & Notes                             \\
\midrule

PHEME \cite{Kochkina2018pheme}                                                      & Twitter         & 105,354 tweets organised into 6425 threads (2402   rumors and 4023 non-rumors), relating to nine events. (A thread consists of tweets introducing a news item and a series of follow-up comments)                                                                                                                       & A & \-                                \\
FakeNewsAMT\cite{perez2018automatic}                                               & various         & 240 fake and 240 legitimate news items                                                                                                                                                                                                                                                                         & A & \-                                \\
Celeb \cite{perez2018automatic}                                                     & various         & 250 fake and 250 legitimate news items  in the celebrity domain                                                                                                                                                                                                                                                                   & A & \-                                \\
Twitter15 \cite{ma2017detectnewtwetter1516}                                         & Twitter         & 1490 source tweets (374 non-rumors, 370 false rumors, 372 true rumors, 374 unverified rumors) with retweets and replies                                                                                                                                                                                                           & A & \-                                \\
Twitter16 \cite{ma2017detectnewtwetter1516}                                         & Twitter         & 818 source tweets (205 non-rumors, 205 false rumors, 205 true rumors, 203 unverified rumors) with retweets and replies                                                                                                                                                                                                            & A & \-                                \\
Twitter16-2\cite{ma2016detectingtwitter16}                                          & Twitter         & 498 rumors and 494 non-rumors with comments                                                                                                                                                                                                                                                         & A & \-                                \\
ISOT \cite{ahmed2018detectingisot,ahmed2017detectionisot}                           & various         & 23481 fake and 21417 genuine news items with titles from 2016-2017, focused on political and world news topics                                                                                                                                                                                                       & A & \-                                \\
LIAR \cite{wang-2017-liar}                                                          & Politifact      & 12.8k manually   labeled short claim statements in various contexts with speaker related meta-data, primarily from 2007-2016                                                                                                                                                                                  & A & \-                                \\
Liar-plus  \cite{alhindi2018your1}                                                  & Politifact      & An extended version of the above LIAR dataset, in which the claims are accompanied by sentences that provide justifications for the assigned labels                                                                                                                                                                                                                                                       & A & \-                                \\
CREDBANK \cite{mitra2015credbank}                                                   & Twitter         & 60 million tweets from 2014-2015, concerning various topics grouped into 1049 real-world events, each labeled by 30 human annotators                                                                                                                                                                                            & A & \-                                \\
Kaggle Fake News dataset \cite{kagglefakenewsdataset2016}                           & various         & 12,999 posts, consisting of both text and metadata, collected over a period of 30 days from 244 websites                                                                                                                                                                                                          & A & \-                                \\
George McIntire dataset  & various         & 6.3k news items, with an equal distribution of fake and real items.  (https://github.com/GeorgeMcIntire/fake\_real\_news\_dataset)                                                                                                                                                                                                                                                               & A & \-                                \\
SLN \cite{rubin2016fakeSLN}                                                         & various         & 360 news articles covering 12 contemporary news topics in 4 domains (civics, science, business, and soft   news)                                                                                                                                                                                                                  & A & \-                                \\
LUN \cite{rashkin2017truthLUN}                                                      & various         & News items classified as trusted(13995), satire(14985), hoax(12047) or propaganda (35029)                                                                                                                                                                                                                                         & A & \-                                \\
Twiter\_harvard\cite{kwon2017rumorTwiter_harvard}                          & Twitter         & 111 events with tweet ids  and user information (60 rumors  and 51 non-rumors)                                                                                                                                                                                                                                                    & A & \-                                \\
health-related news \cite{sicilia2018twitterhealthrelated}                          & Twitter         & 709 posts (54\% rumour, 30\% non-rumour  and 16\% unknown), collected using the keywords \textit{\#zikavirus} and \textit{zika microcephaly}                                                                                                                                                                                                                  & R & \-                                \\
MultiSourceFake\cite{ghanem2021fakeflow2222}                                       & various         & 5,994 real and 5,403 fake news articles                                                                                                                                                                                                                                                                                           & A & \-                                \\
PoliticalNews \cite{castelo2019topicPoliticalNews}                                  & various         & 14,240 news pages from 2013- 2018  (7,136 fake and 7,104 genuine)                                                                                                                                                                                                                                                       & A & \-                                \\
Buzzfeed Political News \cite{horne2017just2}                                       & various         & Dataset 1 (Buzzfeed 2016 election data): 36 real and 35 fake items; Dataset 2 (political news): 75 real, 75 fake and 75 satire items; Dataset 3 (Burfoot and Baldwins satire):  233 satire and 4000 real items                                                                                                                                             & A & \-                                \\
\cite{rezaei2022early22222}                                                         & various         & 23,935 news items from September 1995 to January 2021                                                                                                                                                                                                                                                                                   & A & \-                                \\
HWB \cite{anoop2020emotion158}                                                      & various         & 500 real and 500 fake documents related to health and well being                                                                                                                                                                                                                                                                  & A & \-                                \\
\cite{cuenca2022combining888}                                                       & various         & News articles from eight web sources concerning the Hanoi summit between the presidents of the United States and North Korea,   Donald Trump and Kim Jong-un respectively                                                                                                                                                                                            & N & \-                                \\
MultiFC \cite{augenstein2019multifc}                                                & various         & 36,534 multi-domain claims with their metadata (different domains have different labels, which encompass both direct truth ratings ("correct," "incorrect") and labels that are difficult to map to a level of truthfulness (e.g. ‘grass roots movement!’, ‘misattributed’, ‘not the whole story’)) & A & \-                                \\
Infodemic \cite{patwa2021fightinginfodemic}                                 & various         & 10,700 social media posts and articles (5600 real, 5100 fake) on COVID-19.                                                                                                                                                                                                                                                         & A & COVID-19                          \\
COAID \cite{cui2020coaid}                                                           & various         & 4,251 news items (204 fake and 3,565 true news articles, 28 fake and 454 true claims), 296,000 related user engagements (e.g. clicks, shares), 926 social platform posts about COVID-19.                                                                                                                                        & A & COVID-19                          \\
\cite{COVID-19infodemic2020}                                                        & various         & 586 genuine and 578 fake news items, and more than 1,100 news items and social media posts regarding COVID-19.                                                                                                                                                                                                                     & A & COVID-19                          \\
\cite{arora2022modified232}                                                         & Twitter         & Globally-collected Tweets related to the epidemic, obtained by    filtering tweets containing word or hashtag \textit{Covid-19, Corona Virus, Corona, COVID, covid19}, and \textit{sarscov2}                                                                                                                                            & R & COVID-19                          \\
FakeNewsNet\cite{shu2020fakenewsnet}                                                & Politifact      & 432 fake and 624 real news items with content, images, and social network information                                                                                                                                                                                                                                             & A & multimodal                        \\
FakeNewsNet\cite{shu2020fakenewsnet}                                                & Gossipcop       & 5,323 fake and 16,817  real news items with content, images,   and social network information                                                                                                                                                                                                                                     & A & multimodal                        \\
Fakeddit \cite{nakamura2020fakeddit}                                                & Reddit          & 1,063,106 samples with submission title, image, comments and metadata                                                                                                                                                                                                                                                             & A & multimodal                        \\
MediaEval2016 \cite{inproceedingsMediaEval2016}                                     & Twitter         & 193 cases of real and 220 cases of misused images/videos, associated with 6,225 real and 9,596 fake posts posted by 5,895 and 9,216 unique users, respectively                                                                                                                                                                     & A & multimodal                        \\
NovEmoFake \cite{kumari2023identifying999}                                          & various         & 6816 real and 4950  fake news items (text and images) with background information (where and in which context the news item was first published)                                                                                                                                                                                                                                                  & R & multimodal                        \\
MMM \cite{gupta2022mmm1111}                                                         & various         & 5630  real and 4840 fake news items (text and images) with background information (where and in which context the news item was first published)                                                                                                                                                                                                                                            & A & multimodal,Hindi, Bengali, Tamil \\
Multimodal-Weibo\cite{jin2017multimodalweibo}                                      & Weibo           & 9528 posts (4749 rumor and 4779 non-rumor) with images, created between May 2012 and January 2016                                                                                                                                                                                                                                            & N & multimodal, Chinese               \\
\bottomrule
\end{tabular}
\end{table*}
 

\begin{table*}[htb]
\footnotesize
\begin{tabular}{p{2.3cm}p{0.9cm}p{10cm}p{0.5cm}p{2cm}}
\toprule
Dataset                                                                             & Source          & Description                                                                                                                                                                                                                                                                                                                       & A & Notes                             \\
\midrule
Weibo21 \cite{nan2021mdfendweibo21}                                                 & Weibo           & 4,488 fake and 4,640   real news items from 9 different domains collected between December 2014 and March 2021 with   news text, image content,   timestamps, and comments                                                                                                                                                        & A & multimodal, Chinese               \\
Weibo16 \cite{ma2016detectingtwitter16}                                             & Weibo           & 2313 rumors and 2351 non-rumors with comments                                                                                                                                                                                                                                                                                     & A & Chinese                           \\
Weibo-16 (deduplication) \cite{zhang2021miningwww}                                   & Weibo           & 3706 news items (1,355 fake, 2351 real) with comments                                                                                                                                                                                                                                                                              & A & Chinese                           \\
Weibo-20 \cite{zhang2021miningwww}                                                  & Weibo           & 6362 news items (3161 fake, 3201 real) with comments                                                                                                                                                                                                                                                                               & A & Chinese                           \\
Weibo20-miao \cite{miao2021syntax21}                                                & Weibo           & 3034 rumors and 3034 non-rumors created between 2016 and 2020                                                                                                                                                                                                                                                                                 & A & Chinese                           \\
\cite{guo2019exploiting33}                                                          & Weibo           & 7880 fake and 7907  real news items with approximately 160k comments                                                                                                                                                                                                                                                              & N & Chinese                           \\
Portuguese-data\cite{da2023sentiment666}                                       & various         & 76,782 news items, labeled according to whether they were sourced from true or fake news sites prelabeled                                                                                                                                                                                                                                                                                                    & N & Portuguese                        \\
FakeNewsSet\cite{da2020fakenewssetgen}                                            & Twitter         & 300 fake and 300 genuine news items                                                                                                                                                                                                                                                                                               & A & Portuguese                        \\
Fake.Br \cite{monteiro2018contributionsfakebr}                                      & various         & 3,600 fake and 3,600 genuine news items classified into  six   categories (politics, TV \& celebrities, society \& daily news, science \& technology, economy, and religion)                                                                                                                                                                    & A & Portuguese                        \\
CLEF2020\cite{barron2020overview}                                                        & competi-tion     & Five tasks related to verification of claims: task1: check-worthiness of tweets (962 tweets in English and 7,500 tweets in Arabic); task 2 - verified claim retrieval (1,197 tweets and 10,375 verified claims in English); task 3 - evidence retrieval (200 claims and 14,742 corresponding Web pages containing evidence).; task 4 - claim verification (165 claims in Arabic); Task 5 - check-worthiness on debates (70 debate transcripts in English). All Tasks will run in English. Additionally, tasks 1, 3, and 4 will also run in Arabic and Spanish.           & A & English, Arabic      \\
\cite{al2023exploring}                                                              & Twitter         & 202 false rumors and  201 true rumors relating to 403 real-world events with comments                                                                                                                                                                                                                                             & R & Arabic                            \\
DAST \cite{lillie2019joint99}                                                       & Reddit          & 3007 source posts (273 Support, 300 Deny, 81 Query, 2353 comments); 3007 top-level comments (261 Support, 632 Deny, 304 Query, 1810 comments)                                                                                                                                                                                                & A & stance, Danish                    \\
ByteDance fake news dataset \cite{BytedanceWSDMCup2019}                             & Byte-Dance       & 320,767 news pairs in both Chinese and English;   test data contains 80,126 news pairs. Given the title of a fake news article A and the title of an incoming news article B, participants are asked to classify B according to whether it agrees with A, disagrees with A or is unrelated to A                                                                          & A & stance, Chinese and English       \\
RumourEval17-Task8 \cite{derczynski2017semeval}                                     & Twitter         & Task A (stance classification): 5568 posts (1004 Support, 415 Deny, 464 Query, 3685 comments); Task B (veracity prediction): 325 source posts (145 True, 74 False, 106 Unverified) with associated comments                                                                                                                                                                         & A & stance                            \\
RumourEval-19-Task7 \cite{gorrell2019semeval}                                       & Twitter, Reddit & Task A (stance classification): 8574 posts (1184 Support, 606 Deny, 608 Query, 6176 comments); Task B (veracity prediction): 446 source posts (185 True, 138 False, 123 Unverified) with associated comments                                                                                                                                                                        & A & stance                            \\
FNC-1 \cite{ferreira2016emergent-FNC,fnc12017}                                      & Snopes, Twitter & 49972 tuples,  each consisting of a headline-body pair                                                                                                                                                                                                                                                                            & A & stance                            \\
Covid-Stance \cite{mutlu2020stance}                                                 & Twitter         & 14,374 tweets (2848 Neutral, 4685 Against, 6841 Favor) related to COVID-19                                                                                                                                                                                                                                                         & A & stance                            \\
Emergent \cite{ferreira2016emergent-FNC}                                            & various         & 300 claims, and 2,595 associated article headlines                                                                                                                                                                                                                                                                                & A & stance                            \\
PHEME\_stance \cite{Arkaitz2016PHEMERS}                                             & Twitter         & 297 threads containing 4,561 tweets (including retweets), spanning 138 rumors organised into 9 stories                                                                                                                                                                                                                           & A & stance                            \\
London Riots \cite{lukasik2015classifyinglondondataset}                     & Twitter         & 7297 tweets concerning 7 different rumors (5761 support, 957 deny, 579 question)                                                                                                                                                                                                                                                                   & N & stance                            \\
\cite{giasemidis2016determining}                                                    & Twitter         & 327,484 tweets concerning 72 rumors (60.9\% support, 27.4\% against)                                                                                                                                                                                                                                                                     & N & stance                            \\
2020 US Presidential Election \cite{kawintiranon2021knowledge}                      & Twitter         & 2500 tweets manually labeled with stance, 1250 for each presidential candidate (Joe  Biden and Donald Trump)                                                                                                                                                                                                                      & A & stance                            \\
Sydney Siege \cite{zeng2016unconfirmed}                                         & Twitter         & 4375 tweets (2906 affirm, 1469 deny)                                                                                                                                                                                                                                                                                              & N & stance   \\           \bottomrule
\end{tabular}
\end{table*}

\begin{table*}[htb]
\footnotesize
\caption{Summary of emotion-based misinformation detection. EF: Emotion Features, E: Emotion, S: Sentiment, IE: Image Emotion, ED: Emotion Detection, ERD: Emotion-based Rumor Detection, MLs: Various Machine Learning methods. Other abbreviations are explained in section \ref{sec:TML}.}
\label{tab:methods} 
\begin{tabular}{p{0.3cm}p{0.3cm}p{3.5cm}p{0.5cm}p{2.5cm}p{4cm}p{3.6cm}}

\toprule

Pub                                           & Year             & Data                                                         & EF & ED                                   & ERD                                     & Other Features                                       \\
\midrule

\cite{guo2019exploiting33}                             & 2019       & Customized                                                    & E                                             & GRU                                  & EFN (Fig.~\ref{dual_emotion} (c))                                          & \-                                                   \\
\cite{cui2019sameaa}                                   & 2019       & PolitiFact, GossipCop                                         & S                                             & VADER                                & SAME                                                                   & Image, User-based                                    \\
\cite{bhutani2019fake2}                                & 2019       & Various Public Dataset                                        & S                                             & Naive Bayes                          & NB, RF                                                                 & tf-idf scores, Cosine similarity scores              \\
\cite{ajao2019sentiment23}                             & 2019       & PHEME                                                         & S                                             & LIWC                                 & MLs, LSTM-HAN                                                          & Topics                                               \\
\cite{wang2019rumor2789123}                            & 2019       & Weibo16                                                       & E                                             & ALO                                  & SD-DTS-GRU (Fig.~\ref{Temporal} (b))                                       & Time                                                 \\
\cite{wang2020rumor}                                   & 2020       & Weibo16, Twitter16                                            & S, E                                          & Dictionaries                         & SD-TsDTS-CGRU (Fig.~\ref{Temporal} (b))                                    & Time                                                 \\
\cite{dong2020rumor}                                   & 2020       & PHEME, Twitter15, Twitter16                                   & S                                             & \-                                   & a Hierarchical Attention Network with User and Sentiment   information & User-based                                           \\
\cite{anoop2020emotion158}                             & 2020       & HWB                                                           & E                                             & NRC Intensity                        & MLs                                                                    & \-                                                   \\
\cite{de2020linguisticputaoya}                         & 2020       & Fake.Br                                                       & S, E                                          & Dictionaries                         & MLs                                                                    & Grammatical, Stylistic                               \\
\cite{ding2020fake312}                                 & 2020       & FakeNewsNet, CredBank                                         & S                                             & Dictionaries                         & DT, Bi-LSTM                                                            & Topics                                               \\
\cite{touahri2020evolutionteam222}                     & 2020       & CLEF2020                                                      & S                                             & Dictionaries                         & Web Check                                                              & Topics, Offense Named Entities                       \\
\cite{ezeakunne2020sentiment11111}                     & 2020       & FakeNewsNet                                                   & S                                             & VADER                                & MLs, DNN                                                               & Retweet Rate                                         \\
\cite{wang2021rumor151}                                & 2021       & Weibo16                                                       & S                                             & BERT                                 & TDIE (Fig. ~\ref{Temporal} (a))                                            & Time, Propagation Structure                          \\
\cite{kumari2021multitask3}                            & 2021       & ByteDance, FNC, Covid-Stance                                  & S, E                                          & BERT, LSTM                           & Multitask (Fig.~\ref{Multitask} (b))                                                              & Textual Novelty                                      \\
\cite{zhang2021miningwww}                              & 2021       & RumourEval19, Weibo16, Weibo20                                & S, E                                          & Dictionaries                         & MDE (Fig.~\ref{dual_emotion} (b))                                          & Dual Emotion                                         \\
\cite{maia2021sentiment233}                            & 2021       & FakeNewsSet                                                   & S, E                                          & Dictionaries                         & MLs                                                                    & Image Captioning, Grammatical, Stylistic             \\
\cite{miao2021syntax21}                                & 2021       & Weibo16, Weibo20                                              & S                                             & ALO, BERT                            & SSE-BERT (Fig.~\ref{detectwithtree} (a))                                   & Dependency Tree                                      \\
\cite{ghanem2021fakeflow2222}                          & 2021       & MultiSourceFake, LUN, PoliticalNews, FakeNewsNet & S, E                                          & Bi-GRUs basd on dictionary           & FakeFlow (Fig.~\ref{textbasedfeatures} (a))                                  & Topics                                               \\
\cite{choudhry2022emotion}                             & 2022       & PHEME, FakeNews AMT, Celeb, Gossipcop                         & E                                             & Unison model                         & Multitask (Fig.~\ref{Multitask} (a))                                       & Domains                                              \\
\cite{kolev2022foreal321}                              & 2022       & ISOT                                                          & E                                             & RoBERTa                              & RoBERTa, RF                                                            & \-                                                   \\
\cite{iwendi2022covid11}                               & 2022       & \cite{COVID-19infodemic2020}                                  & S                                             & \-                                   & GRU, LSTM, RNN                                                         & Stylistic, Linguistic-informed                       \\
\cite{dong2022sentiment1hh}                            & 2022       & Twitter15, Twitter16                                          & S                                             & RoBERTa                              & SA-HyperGAT (Fig.~\ref{Graph-based} (b))                                   & Structure                                            \\
\cite{arora2022modified232}                            & 2022       & Customized                                                    & S                                             & VADER                                & Modified VADER                                                         & Diffused Information                                 \\
\cite{kelk2022automatic444}                            & 2022       & MultiFC                                                       & S, E                                          & EmoAttention BERT                    & EmoAttention BERT (Fig.~\ref{textbasedfeatures} (c))                & \-                                                   \\
\cite{mohamed2022applying777}                          & 2022       & Infodemic, CoAID                                              & S                                             & Fuzzy Sentiment Scoring              & LSTM (with Fuzzy Sentiment) (Fig.   ~\ref{textbasedfeatures} (d))   & \-                                                   \\
\cite{cuenca2022combining888}                          & 2022       & Customized                                                    & S                                             & SenticNet                            & Conceptual Graphs with sentiment                                       & Entity                                               \\
\cite{gupta2022mmm1111}                                & 2022       & MMM \cite{kumari2023identifying999}                           & IE                                            & Resnet18                             & SCL, BERT and ResNET18                                                 & Novelty, Image,                                      \\
\cite{haque2022graph3333}                              & 2022       & PHEME                                                         & S, E                                          & NRC                                  & GCS (Fig.~\ref{Graph-based} (c))                                           & N-gram, Similarity Matching                          \\
\cite{uppada2022novel4444}                             & 2022       & MediaEval2016, Multimodal-weibo                               & IE                                            & VGG16                                & CredNN                                                                 & \-                                                   \\
\cite{uppada2022image5555}                             & 2022       & Fakeddit                                                      & IE                                            & Xception                             & BERT, Xception                                                         & Image, image caption                                 \\
\cite{kumari2022fake6666}                              & 2022       & ByteDance, Covid-Stance, FNC, LIAR-PLUS                       & E                                             & BERT                                 & LR                                                                     & Novelty                                              \\
\cite{fu2022rumor7777}                                 & 2022       & Weibo16, RumourEval                                           & S                                             & Dictionaries                         & RvNN with Temporal (Fig.~\ref{Temporal} (c))                               & Time                                                 \\
\cite{seddari2022hybrid9999}                           & 2022       & Buzzfeed Political News                                       & S                                             & SEO Scout’s analysis                 & MLs                                                                    & Stylistic, Linguistic-informed                       \\
\cite{rezaei2022early22222}                            & 2022       & Customized                                                    & S                                             & Afinn, VADER                         & MLs                                                                    & Topics, Title-text similarity                        \\
\cite{al2023exploring}                                 & 2023       & Customized                                                    & S, E                                          & Dictionaries                         & Arabic PLMs                                                            & {\color[HTML]{222222} Topics, Stylistic, User-based} \\
\cite{luvembe2023dual14}                               & 2023       & RumourEval19, PHEME, Fakeddit                                 & S, E                                          & CNN, Bi-GRU                          & AGWu-RF (Fig.~\ref{dual_emotion} (a))                                      & Stylistic                                            \\
\cite{balshetwar2023fake21}                            & 2023       & ISOT, LIAR                                                    & S                                             & Lexicon-based                        & MLs                                                                    & Grammatical, Likes                                   \\

\cite{hamed2023fake46}                                 & 2023       & Fakeddit                                                      & S, E                                          & Dictionaries                         & Bi-LSTM                                                                & Title                                                \\
\cite{guo2023tiefake29}                                & 2023       & PolitiFact, GossipCop                                         & S, E                                          & method in \cite{zhang2021miningwww}  & BERT, ResNeSt-50                                                       & Title-Text similarity,Images                         \\
\cite{fang2023unsupervised11}                          & 2023       & Weibo16, Twitter15, Twitter16                                 & S                                             & Sentiment Pattern Module (SPM)       & ptVAE (Fig.~\ref{detectwithtree} (b))                                      & User-based, Structure, Propagation                   \\
\cite{zhang2023sentence333}                            & 2023       & SLN, LUN                                                      & S, E                                          & RoBERTa, Sentiment Interaction Graph & MHN (Fig.~\ref{Graph-based} (a))                                           & Graph, Topics, Entities                              \\
\cite{ali2023rumour555}                                & 2023       & Twiter-harvard,    health-related                             & E                                             & NRC                                  & MLs                                                                    & Linguistic-informed, User-based                      \\
\cite{da2023sentiment666}                              & 2023       & Portuguese datsaset (custumized), Fake.br                     & S                                             & Sentiment Gradient                   & MLs, LSTM                                                              & \-                                                   \\

\bottomrule
\end{tabular}
\end{table*}
 

\begin{table*}[htb]
\footnotesize
\begin{tabular}{p{0.3cm}p{0.3cm}p{3.5cm}p{0.5cm}p{2.5cm}p{4cm}p{3.6cm}}
\toprule
Pub                                           & Year             & Data                                                         & EF & ED                                   & ERD                                     & Other Features                                       \\
\midrule

\cite{kumari2023identifying999}                        & 2023       & NovEmoFake (custumized)                                       & IE                                            & Resnet18                             & SCL, BERT and ResNET18, VisualBert                                     & Novelty, Image                                       \\
\cite{zhao2023collaborative8888}                       & 2023       & Weibo21                                                       & S                                             & BERT                                 & Mixture-of-Experts (Fig.~\ref{textbasedfeatures} (b))                        & \-                                                   \\
\cite{zhang2023sentiment33333}                         & 2023       & Weibo16                                                       & S                                             & Bi-GRU with Attention model          & Bi-GCN                                                                 & Semantic, Propagation information                    \\
\cite{pillai2023misinformation44444}                   & 2023       & \cite{barbieri2020tweeteval}                                  & S, E                                          & RoBERTa, DistilBERT                  & An Ensemble Method, RNN                                                & \-                                                   \\
\cite{choudhry2022emotion33,chakraborty2023emotion233} & 2022, 2023 & FakeNewsAMT, Celeb, Politifact and Gossipcop datasets         & E                                             & Unison model                         & Multitask (Fig.~\ref{Multitask} (a))                                       & Domains            \\
\bottomrule
\end{tabular}
\end{table*}

\subsection{Conventional Machine Learning Methods \label{sec:TML}}

Machine learning is a branch of AI that uses algorithms and statistical models to teach computers how to make predictions and decisions automatically. As shown in Table~\ref{tab:methods}, a variety of machine learning algorithms has been used to develop misinformation classifiers. These include both supervised methods, such as passive-aggressive \cite{balshetwar2023fake21}, Naive Bayes (NB) \cite{bhutani2019fake2}, k-Nearest Neighbour (KNN) \cite{da2023sentiment666,maia2021sentiment233}, Support Vector Machine (SVM) \cite{ezeakunne2020sentiment11111}, Random Forests (RF) \cite{kolev2022foreal321}, Decision Tree (DT) \cite{ding2020fake312}, AdaBoost (AB) \cite{rezaei2022early22222}, LOGIT, Grad Boosting, XG-Boost, Gradient Boost (GB) \cite{ajao2019sentiment23}, and unsupervised methods like K-Means and DBSCAN \cite{anoop2020emotion158}.

\subsection{Deep Learning Methods}

DL is a sub-field of machine learning that has made breakthrough progress in many fields, especially in computer vision, NLP, speech recognition, and other AI fields \cite{sharifani2023machine6}. Compared to conventional machine learning methods, DL techniques can handle larger and more complex datasets and can result in improved performance on certain tasks \cite{al2023deep2}. DL algorithms build complex models by stacking multiple neural network layers, which are called deep neural networks. Pre-training is a DL model training strategy, in which models are initially trained on a large-scale data set to learn a common feature representation that is suitable for application in a range of different scenarios. Pre-trained models are subsequently fine-tuned to achieve optimal results when applied to specific tasks. As shown in Table~\ref{tab:methods}, DL approaches have been widely used in both sentiment/emotion analysis and misinformation detection. 
For example, Iwendi et al. \cite{iwendi2022covid11} explored the use of RNN, GRU, and LSTM as classifiers to detect fake news relating to COVID-19 based on 39 features (including sentiment, linguistic, and named entities) extracted from news articles and social media posts. 
Ajao et al. \cite{ajao2019sentiment23} employed various machine learning methods and an LSTM with hierarchical attention networks (HAN) \cite{yang2016hierarchical} for rumor detection. A Bi-LSTM was used by Hamed et al. \cite{hamed2023fake46} to detect misinformation using dual emotions and content features. 
In \cite{luvembe2023dual14}, the authors adopted CNN and Bi-GRU to extract dual emotion features. 
To evaluate the effectiveness of their proposed multi-tasking framework for rumor detection, Choudhry et al. \cite{choudhry2022emotion} employed various DL methods, including LSTM, BERT, CNN, RoBERTa, CapsuleNet \cite{sabour2017dynamic3} and HAN.
Various studies have applied GCN and GNN to model the graph-like structure of social media posts \cite{dong2022sentiment1hh,zhang2023sentiment33333}.

Pre-trained models, including BERT, DistilBERT, and RoBERTa, have frequently been used used as the basis for extracting sentiment and emotion features in the context of misinformation detection 
\cite{wang2021rumor151,kolev2022foreal321,kumari2021multitask3,dong2022sentiment1hh,kumari2022fake6666,zhang2023sentence333,zhao2023collaborative8888,pillai2023misinformation44444}. A popular technique has been to use transfer learning to fine-tune these pre-trained models on large emotion detection datasets (e.g., GoEmotions \cite{demszky2020goemotions} and DailyDialogue \cite{li2017dailydialog4}) prior to labeling misinformation datasets.
Moreover, there exists a small number of pre-trained models for languages other than English, such as AraBERT-Twitter and MARBERT, which were used for rumor detection in Arabic social media \cite{al2023exploring}. 

For multimodal data, Resnet18, VGG16 and Xception have been used to extract image features \cite{kumari2023identifying999,gupta2022mmm1111,uppada2022novel4444,uppada2022image5555}. Similarly to text-only datasets, these studies use transfer learning to fine-tune pre-trained image models on visual sentiment datasets, then extract image features from the misinformation dataset, and finally merge them with text features for misinformation detection.

\subsection{Advanced Fusion Methods \label{sec:methodtoemo}}

A wide variety of methodologies for emotion-based misinformation has been developed (See Table \ref{tab:methods} for a complete list). In the majority of cases, information about emotions and/or sentiment is fused with other types of features, aiming to take full advantage of the specific characteristics of the dataset used to maximize detection performance. Additional features may be based, for example, on various aspects of textual content; information regarding the structure or temporality of collections of social media posts; and/or images associated with textual data.  Moreover, approaches vary in terms of whether they carry out learning within the context of single or multi-task framework.  In this section, we introduce a selection of these advanced fusion methods, which are categorized according to the types of additional features and/or the learning strategy that they employ.

\subsubsection{Methods Combining Emotion with Other Text-Based Features}

Various methods have attempted to exploit the wealth of valuable information conveyed in text by combining emotion/sentiment features with other features derived from the textual content of news articles or social media posts.  

\begin{figure*}
\centering
\includegraphics[width=1.8\columnwidth]{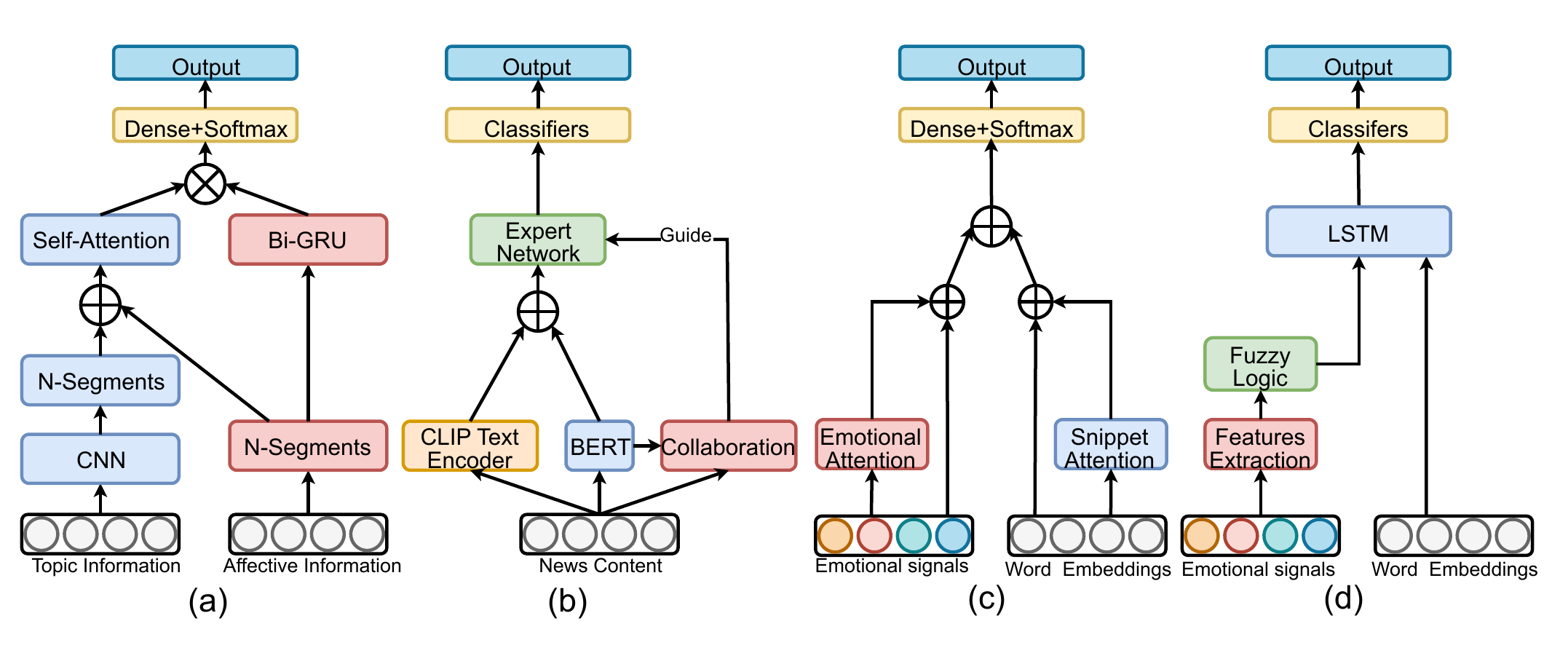}
\caption{Emotion-based misinformation detection by combining emotion with other text-based features. (a) FakeFlow \cite{ghanem2021fakeflow2222}, (b) Mixture-of-Experts\cite{zhao2023collaborative8888}, (c) EmoAttentionBERT\cite{kelk2022automatic444} (d) LSTM (with Fuzzy Sentiment)\cite{mohamed2022applying777}}
\label{textbasedfeatures}
\end{figure*}

Ghanem et al. \cite{ghanem2021fakeflow2222} proposed the FakeFlow model (Figure~\ref{textbasedfeatures} (a)), which aims to model the flow of affective information in fake news articles, based on the hypothesis that the pattern of affective information in fake news differs from that found in genuine news, e.g., emotions of fear are often evoked towards the start of fake new articles. The model consists of two modules, the first encoding topic information, extracted using a CNN, and the second capturing 23 affective features relating to emotion, sentiment, morality, imageability and hyperbola. In the first module, potential relationships between topics and affective information are captured by concatenating their respective vectors  (e.g., emotions in a fake article about Islam are likely to vary from those in an article in favor of a politician). A context-aware self-attention mechanism is subsequently applied to weight segments according to their similarity to neighboring segments. In the second module, the flow of the affective information within the articles is modeled by feeding the affective vectors to Bi-GRUs. Finally, a dot product and average operation are applied to distill the output of the two modules into a compact representation, which is fed into a fully connected layer and a softmax layer to determine the factuality of the article.        

The multi-domain fake-news detection system described in  \cite{zhao2023collaborative8888} (Figure~\ref{textbasedfeatures} (b)) is based on mixture-of-experts model, which involves training multiple neural networks based on TextCNNs (experts), each targeted at a different part (domain) of a dataset.  Pre-trained BERT and CLIP  \cite{radford2021learning1} text encoders are applied to obtain two different embeddings of news content, which are combined as a fusion embedding.  The use of the CLIP text encoder, which is pre-trained on image-text paired datasets, aims to take advantage of the rich semantic representations obtained through state-of-the-art multimodal learning. A \textit{Collaboration} module adaptively determines the weights of each expert model, to enhance or suppress their contribution in the final mixture-of-experts model. The module consists of a fusion vector  C${_i}$, which combines sentence-level embeddings e${_a}$ from attention, sentiment embeddings e${_s}$ obtained by fine-tuning BERT using the Weibo\_senti\_100k dataset, and domain embeddings e${_d}$. The expert networks are accumulated and multiplied via the collaborative influence function C${_i}$, which is determined by the Collaboration module, and then used for classification.

The EmoAttention BERT model architecture \cite{kelk2022automatic444} (Figure~\ref{textbasedfeatures} (c)) uses both emotion and snippet attention to verify the truth of political claims, supported by evidence from Google news snippets. The content of snippets is encoded using word embeddings, while the NRC Intensity Emotion Lexicon \cite{LREC18-AILIntensities} is used to calculate word–level intensities for eight basic emotions. An emotional attention layer assigns a weight to each emotion vector to identify the most relevant emotional signals in a given evidence snippet, while a snippet attention layer weights each evidence snippet with respect to the associated claim. Finally, the vectors from both layers are distilled and fed into a softmax layer to predict the truth of the claim. 

Mohamed et al. \cite{mohamed2022applying777}  detects fake news using an LSTM that combines textual embeddings from Word2Vec with fuzzy sentiment features (Figure ~\ref{textbasedfeatures} (d)). Sentiment features are extracted by firstly identifying opinion-denoting words and associated polarity information using the SentiWordNet \cite{esuli2007sentiwordnet} and WordNet\footnote{https://wordnet.princeton.edu/}, resulting in an initial score. These values are subsequently modified using fuzzy logic functions, according to the presence of different types of linguistic hedges (i.e., words that modify the intensity and meaning of an expressed opinion, such as \textit{not, very,} and \textit{quite}), using fuzzy logic functions to obtain the final sentiment score.

\subsubsection{Mining of Dual Emotions \label{sec:dualemotion}}

A number of studies has investigated how misinformation detection in social media can be improved by taking into account information about the different emotions expressed in posts that announce news (i.e., \textit{publisher posts}) and posts that comment on or react to these source posts (i.e., \textit{social posts})

\begin{figure*}
\centering
\includegraphics[width=1.8\columnwidth]{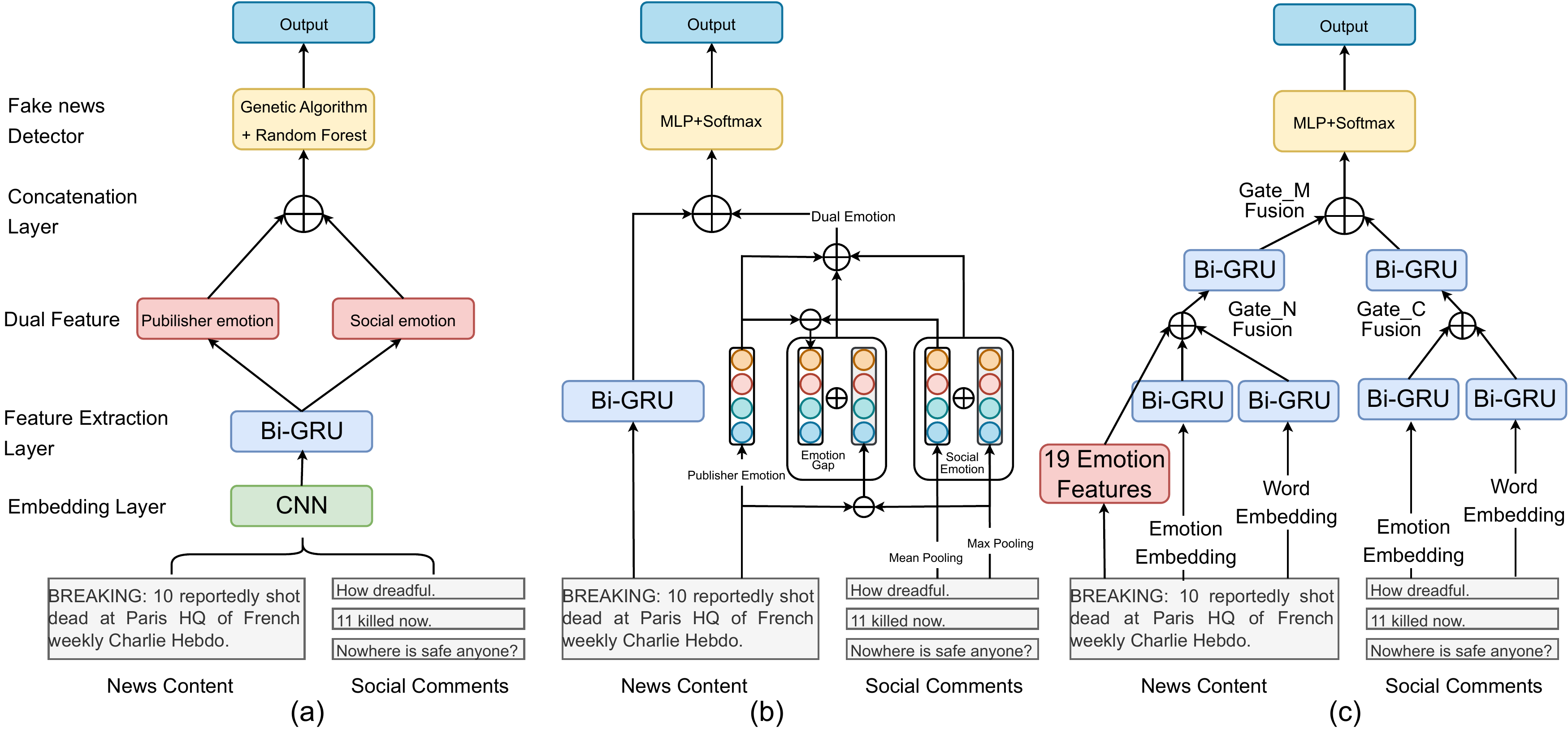}
\caption{Emotion-based misinformation detection by mining of dual emotions. (a) AGWu-RF\cite{luvembe2023dual14}, (b) MDE\cite{zhang2021miningwww}, (c) EFN\cite{guo2019exploiting33}}
\label{dual_emotion}
\end{figure*}

Luvembe et al. \cite{luvembe2023dual14} developed a deep normalized attention-based mechanism for enriched extraction of dual emotion features (Figure~\ref{dual_emotion} (a)), which combines CNN and Bi-GRU. The CNN layer is used to obtain embeddings for both publisher and social posts, after which a stacked Bi-GRU with attention is utilized to extract and concatenate emotion features from each type of post.  Classification of publisher tweets according to whether or not they report misinformation is performed using a random forest model, whose features are guided by a genetic algorithm, which determines the subset of features that can achieve optimal classification performance.

The MDE model \cite{zhang2021miningwww}  (Figure~\ref{dual_emotion} (b)) detects misinformation in social media posts by integrating features from existing Bi-GRU fake news detectors with publisher and social emotion features and the relationship between them. A vector representing emotions in the publisher post emo${_P}$, is obtained by concatenating the emotion category, lexicon-based emotion score, emotional intensity, sentiment score, and other auxiliary features (e.g., emoticons and punctuation). A vector is created for each social post, by applying the same method used to obtain the publisher emotion vector. The individual social emotion vectors are subsequently combined, after which they are aggregated in two ways, i.e. using mean pooling emo${_S^{mean}}$ (to represent average emotion signals) and max pooling emo${_S^{max}}$ (to capture extreme emotional signals). These two types of aggregation are then concatenated to obtain the overall \textit{Social Emotion} emo${_S}$. The \textit{Emotion Gap} emo${_{gap}}$ represents the difference between the publisher and social emotions, and is obtained by concatenating (emo${_P}$ - emo${_S^{mean}}$) and (emo${_P}$ - emo${_S^{max}}$). Finally, dual emotion features are obtained by concatenating the publisher emotion (emo${_P}$), the social emotion (emo${_S}$) and the Emotion Gap (emo${_{gap}}$). These features are combined with those from the existing Bi-GRU fake news detector, and fed into a multi-layer perceptron (MLP) layer and a softmax layer to determine whether or not the publisher post represents fake news.  

The end-to-end emotion-based fake news detection framework for social media (EFN) proposed by Guo et al. \cite{guo2019exploiting33} (Figure~\ref{dual_emotion} (c)) consists of a content module, comment module and fake news detection module. The content module (left of the figure) is used to encode publisher posts using Bi-GRUs for word embeddings and emotion embeddings, the latter of which is trained using large-scale Weibo datasets, with emoticons as the emotion labels. A gate recurrent unit (Gate\_N) is then applied to combine word embeddings, emotion embeddings, and 19 sentence-based emotion features. Subsequently, all vectors are fed into another Bi-GRU, whose final hidden state is used as the representation of the publisher post. The comment module (right of the figure)  represents information about follow-up social posts. The comment module architecture is similar to the content module, except that all comments are concatenated before being fed into the Bi-GRU, and sentence-based emotion features are not used. A different Gate (Gate\_C) is used to fuse features. Finally, the output of the third Gate (Gate\_M), which combines the content and comment representations, is fed to a fully connected layer with softmax activation to determine whether or not the publisher post constitutes fake news.

\subsubsection{Methods Based on Tree or Graph Structures \label{sec:treegraph}}

Due to the inherent relationships among posts relating to fake news, such as retweets or likes of source posts from followers on Twitter, social media data may be viewed as tree structures and graph structures through which information propagates. Accordingly, several methods employ tree or graph structures to model the spread of information and capture the relationship between nodes of the tree. The words and phrases that make up sentences can also be arranged into hierarchical tree-like structures, according to the grammatical and semantic relationships that hold between them. Several misinformation detection methods make use of features based on these relationships, including dependency tree and sentiment tree information. 

\begin{figure*}
\centering
\includegraphics[width=1.8\columnwidth]{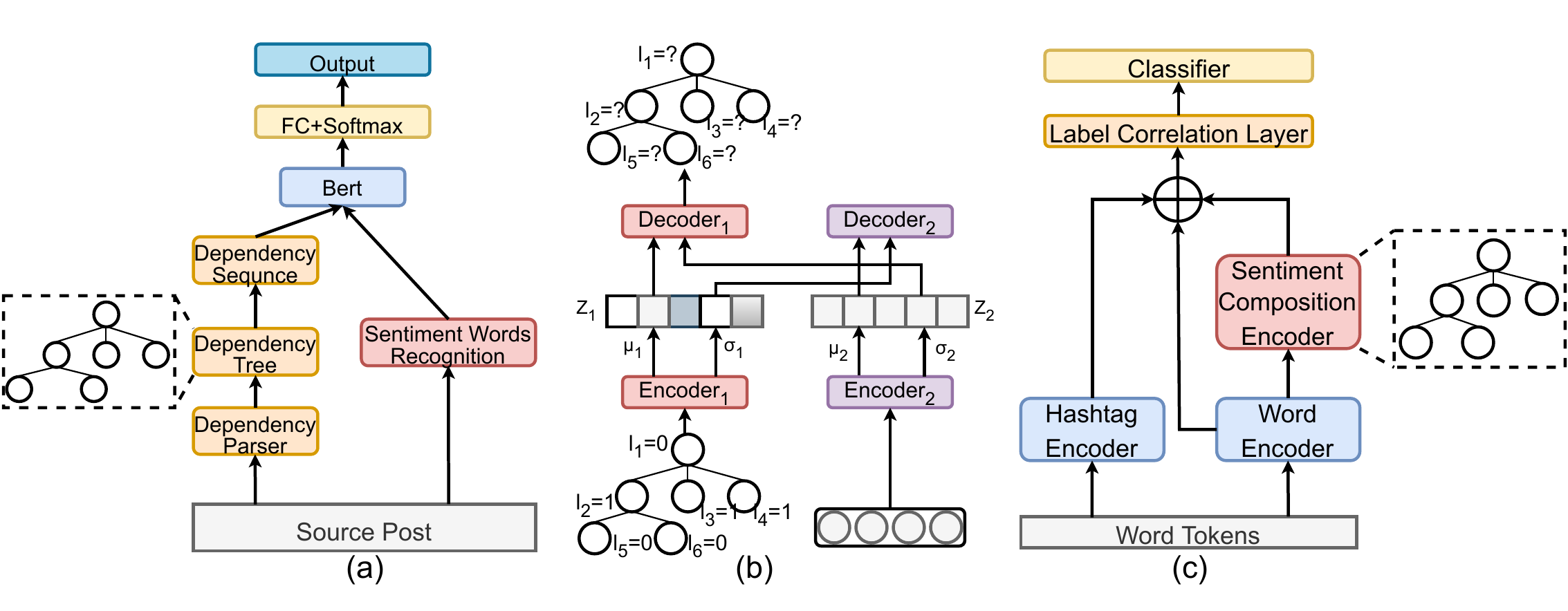}
\caption{Emotion-based misinformation detection based on tree structures. (a) SSE-BERT\cite{miao2021syntax21}, (b) ptVAE \cite{fang2023unsupervised11}, (c) Multi-EmoBERT\cite{li2023multi2223}}
\label{detectwithtree}
\end{figure*}

In \cite{miao2021syntax21}, an \textit{earliest rumor detection} approach for social media is described (Figure~\ref{detectwithtree} (a)). It considers only publisher posts without their follow-up social comments, with the aim of catching potentially harmful rumors before they become widespread. The use of a syntax and sentiment enhanced version BERT (SSE-BERT) is inspired by the observations that both the sentiment \textit{and} syntactic features of rumors are often distinct from non-rumors.  Syntactic dependency trees firstly are obtained for each source post using DDParser \cite{zhang2020practical0914}, and are encoded into a dependency sequence by preorder traversal. Sentiment-denoting words in seven different categories are then recognised using an external sentiment lexicon (i.e., ALO \cite{xu2008constructingALO}). Specific embeddings are then assigned to each token according to the sentiment lexicon. All features are learned by BERT and distilled using element-wise addition. Finally, the vector of [CLS] in BERT is fed into a fully connected network with softmax to detect rumorous publisher posts.

Driven by the scarcity of high-quality annotated training data, \cite{fang2023unsupervised11} developed an unsupervised approach, ptVAE (Figure~\ref{detectwithtree} (b)). Based on the observations that rumorous tweets exhibit different sentiment patterns compared to rumorous tweets, and that they diffuse more rapidly, deeply and broadly, the method aims to detect rumors by identifying collections of tweets whose propagation patterns and sentiment characteristics differ from those of normal (i.e, non-rumorous) collections. The proposed model consists of a Sentiment Pattern Module (SPM), Propagation Feature Module (PFM), and Cross-alignment module. In the SPM (left of Figure~\ref{detectwithtree} (b)), a tree encoder infers the pattern of sentiment labels along the input propagation tree and uses a GRU to encode this pattern into a latent vector z${_{1}}$. The original sentiment labels for each node are then reconstructed by decoding z${_{1}}$ using a node label decoder and a child label distribution decoder which, respectively, predict the label of each node and the label distribution of the node's children. The PFM (right of Figure~\ref{detectwithtree} (b)) creates vectors capturing the speed, and depth \& breadth of propagation, and combines them as the input to a VAE, whose encoder and decoder are based on a multilayer perception. The Cross-alignment module then jointly learns the propagation tree of the SPM and the propagation characteristics of the PFM.     

Li et al. \cite{li2023multi2223} propose the Multi-EmoBERT model (Figure~\ref{detectwithtree} (c))  to detect multiple co-existing emotions in fake news content on social media platforms. The first part consists of a Word Encoder to obtain representations of words, and a Hashtag Encoder to obtain representations of emotion-word hashtags and emojis, which are common features of social media text. The second part is the Sentiment Semantic Composition Encoder, which uses the Stanford CoreNLP toolkit to construct sentiment trees, and employs a self-attention mechanism and phrase node selection to obtain phrase level vectors. The final part is a label correlation layer that uses a parameter to capture correlations between co-existing emotions. A subsequent analysis, revealing that multiple emotions are often conveyed within a single fake news posting, demonstrates the potential value of Multi-EmoBERT in detecting fake news posts.

\begin{figure*}
\centering
\includegraphics[width=1.8\columnwidth]{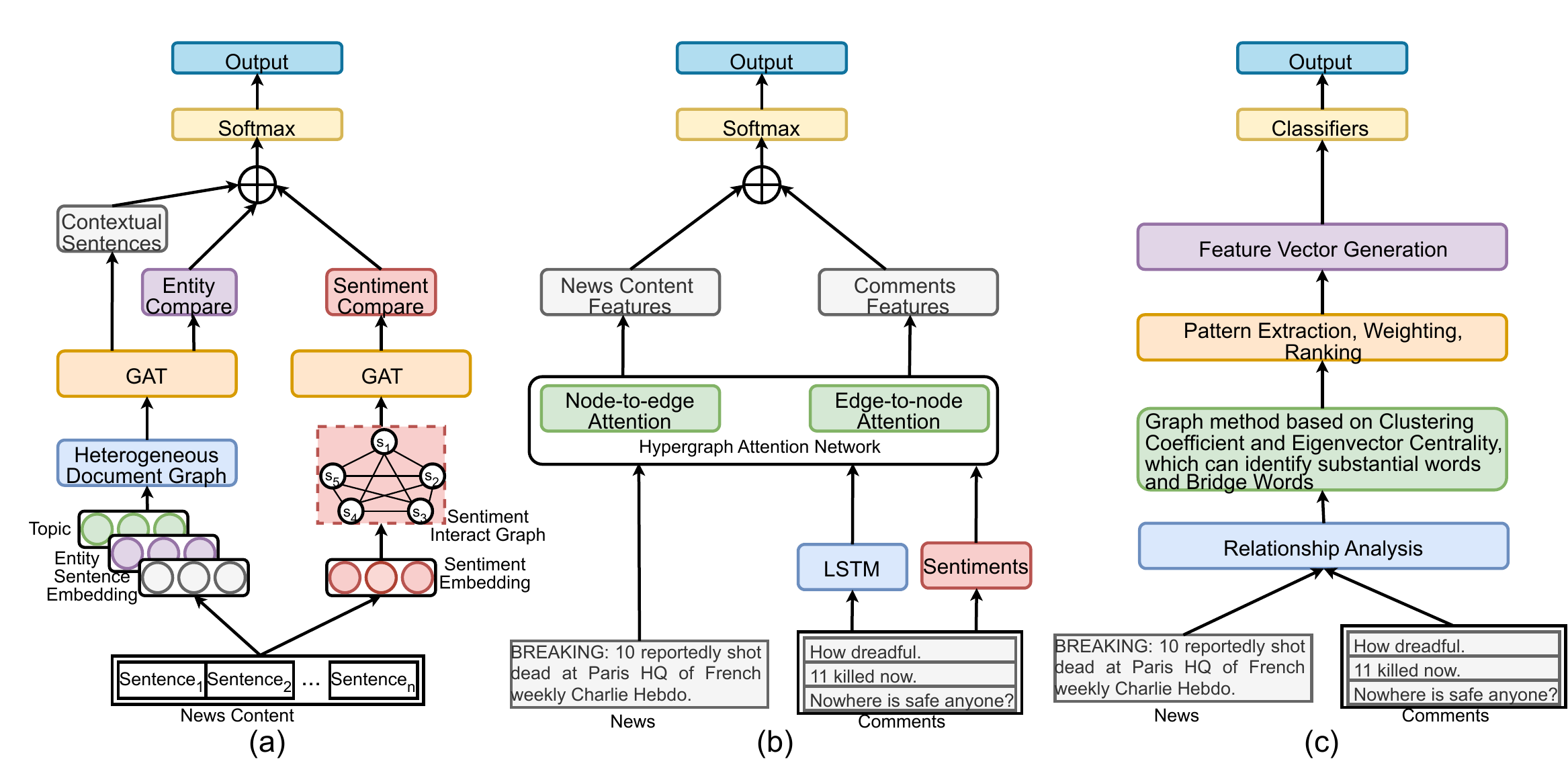}
\caption{Emotion-based misinformation detection based on graph structure. (a) MHN\cite{zhang2023sentence333}, (b) SA-HyperGAT \cite{dong2022sentiment1hh}, (c) GCS\cite{haque2022graph3333}}
\label{Graph-based}
\end{figure*}

The method described in \cite{zhang2023sentiment33333} combines the use of semantic and sentiment information, along with the structure of information propagation in social network posts to obtain enriched features for rumor detection. BERT is used to separately capture information about publisher tweets and follow-up social comments, while Bi-GRU with Attention is used to encode sentiment information conveyed in follow-up tweets.  Propagation features of tweets are obtained with the aid of a Bi-GCN network. The various features are then spliced and fused to detect rumors. 

Figure~\ref{Graph-based} (a) illustrates the graph attention network-based model (MHN) developed by Zhang et al. \cite{zhang2023sentence333}, aimed at detecting longer news articles that contain misinformation. The approach is based on the finding that patterns of sentiments expressed across sentences in fake news articles are usually very different from patterns in real news articles. The model employs two types of graph structures. Firstly, a sentiment interaction network encodes sentence-level sentiment features using a pre-trained RoBERTa model, and captures changes in sentiment in that occur in the context of the surrounding sentences. A sentiment comparison model calculates comparison vectors between each contextual sentiment representation obtained from the input news document and its corresponding original sentiment embedding; discrepancies between these embeddings could be indicative of fake news. Secondly, a heterogeneous document graph encodes the semantic content of the article, by capturing interactions between sentences, topics, and entities. A comparison between the contextual entity vectors and those obtained from a knowledge graph is aimed at detecting potential information inconsistencies that could denote fake news. The sentiment comparison vector, entity comparison vector and article representations are combined and passed through a Softmax layer to make predictions.

Dong et al. \cite{dong2022sentiment1hh} designed a Sentiment-Aware Hypergraph Attention Network (SA-HyperGAT) for fake news detection in social media (Figure~\ref{Graph-based} (b)). The use of hypergraphs is intended to capture higher-order dependency information between words and sentences, compared to general graphs. Separate hypergraphs are constructed for publisher posts and follow-up social comments. In the former hypergraph, each node corresponds to a word in the news text, while in the latter, nodes correspond to user comments. Sentiment labels for each comment, obtained using a fine-tuned RoBERTa model, are used as hyperedges in the graph. Representations of comments are learnt using an LSTM, after which node-to-edge attention and edge-to-node attention are applied to learn the representation of the hypergraphs. Final feature vectors are obtained by applying mean pooling to both hypergraphs; these vectors are combined and then fed into a softmax classifier to obtain the final prediction.

The graph-based contextual and semantic learning (GCS) method for detecting rumors in tweets \cite{haque2022graph3333} (Figure~\ref{Graph-based} (c)) is based on a novel approach to graph-based representation learning, and the identification of two prevalent categories of words that constitute the building blocks for constructing contextual patterns for rumor detection, i.e., \textit{substantial words}, which are used to express emotions, sentiments, or suspicions about the event, and \textit{bridge words}, which connect substantial words.   After data pre-processing, publisher and social tweets are combined to allow important relationships to be identified, e.g., social tweets may convey skepticism, correction, verification, etc, towards the publisher tweet. The combined tweets are represented as word co-occurrence graphs, to which \textit{clustering coefficient} and \textit{eigenvector centrality} are applied to identify substantial and topical words and bridge words, respectively. These are further enriched with negative emotional patterns and skeptical patterns. Next, a modified TF-IDF formula is used to rank and select the top-k patterns most likely to be indicative of rumor.  Semantic vectors are then generated for both tweets and patterns using word embeddings, which are combined and then converted into features using cosine similarity for subsequent use by different conventional machine classification algorithms (i.e., support vector machine, gradient boosting, conditional random field, and logistic regression).    

\subsubsection{Methods Based on Temporal Information\label{sec:timeinfo}}

Various temporal features have been explored to enhance the performance of misinformation detection, based on the time-sensitive patterns that are frequently observed in social media, e.g., rumors initially spread quickly but gradually disappear, while reader emotions tend to change over time. 

\begin{figure*}[!t]
\centering
\includegraphics[width=2\columnwidth]{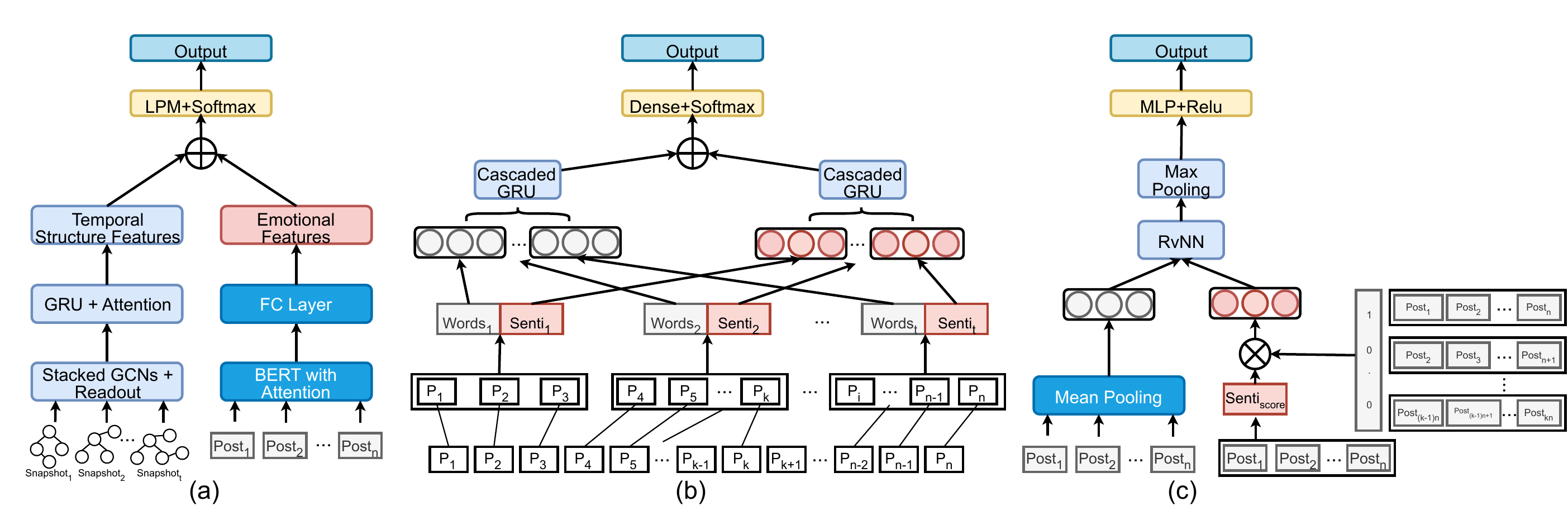}
\caption{Emotion-based misinformation detection based on temporal information. (a) TDEI\cite{wang2021rumor151}, (b) SD-TsDTS-CGRU\cite{wang2019rumor2789123,wang2020rumor}, (c) RvNN with Temporal \cite{fu2022rumor7777}}
\label{Temporal}
\end{figure*}

The TDEI model \cite{wang2021rumor151} (Figure~\ref{Temporal} (a)) integrates emotion features with information concerning time-sensitive dynamic changes in the topological propagation structure of tweets, which is considered to be a better predictor of rumors than the final, static propagation structure.  The graph representing the propagation structure of a publisher post and its associated social comments is firstly divided into a sequence of temporal snapshot graphs.  Stacked GCNs and a readout function are used to learn structural features of the snapshots. A GRU with self-attention is then applied to learn the diffusion process of structures. Meanwhile, emotion vectors are extracted from each post using a pre-trained, fine-tuned BERT model. A self-attention mechanism is then used to merge the emotion vectors for each post corresponding to a rumor event into a single vector, whose dimensionality is adjusted using a fully connected layer.  The temporal dynamic structure and emotion vector are then concatenated and fed into multi-layer perception with softmax function to make predictions.

 The SD-TsDTS-CGRU fusion rumor detection method \cite{wang2019rumor2789123, wang2020rumor} (Figure~\ref{Temporal} (b)) focuses on detecting rumors at the event level, i.e., by considering all information expressed in the complete set of sequential posts related to the same topic or event.  Posts are firstly automatically partitioned into sets covering distinct events by dividing them into intervals using a two-step dynamic time series division algorithm, based on fuzzy clustering and information granules \cite{zadeh1979advances2}. The latter step helps to ensure that each batch of posts covers information at an appropriate level of granularity and has a consistent semantic interpretation.  The calculation of information granularity takes into account the number of sentiment words belonging to each fine-grained sentiment category in each interval, obtained using a novel sentiment dictionary containing sentiment words and emoticons. Following the division, word embeddings and sentiment information extracted from the posts in each event-specific set are fed to two different GRUs, whose outputs are combined and fed into a dense layer with Softmax function to predict whether or not each set of event-related posts constitutes a rumor.

Temporal sentiment features of rumors are employed in \cite{fu2022rumor7777} (Figure~\ref{Temporal} (c)) to account for changes in sentiment over the lifetime of an original publisher post and its associated social posts in both Chinese and English social media datasets. The Baidu sentiment API\footnote{https://ai.baidu.com/tech/nlp\_apply/sentiment\_classify} and
NLTK sentiment module\footnote{https://www.nltk.org/api/nltk.sentiment.html?highlight=sentiment\#\\module-nltk.sentiment} are used to obtain sentiment scores for Chinese and English posts, respectively. Temporal features are characterized using a one-hot vector, whose length is modified by normalizing the number of posts in the reply series. Temporal sentiment features are obtained by multiplying the modified one-hot vector with the sentiment score. Textual features of posts are obtained using pre-trained word embeddings and a mean pooling layer, which are combined with the sentiment vector to derive a microblog representation. An RvNN and max pooling layer are then used to obtain a comprehensive representation of an event as it propagates through the path of social replies, which is passed to a MLP with ReLU to determine whether or not the publisher post is a rumor.

\subsubsection{Multitask Learning \label{sec:multitask}}

Multi-task learning optimizes several learning tasks simultaneously, exploiting shared information to improve the prediction performance of the model for each task. Auxiliary tasks can be added to the main task to boost the performance. Several studies have explored how emotion and sentiment detection can act as auxiliary tasks in a multi-task learning framework to enhance misinformation detection accuracy.

\begin{figure}[!t]
\centering
\includegraphics[width=\columnwidth]{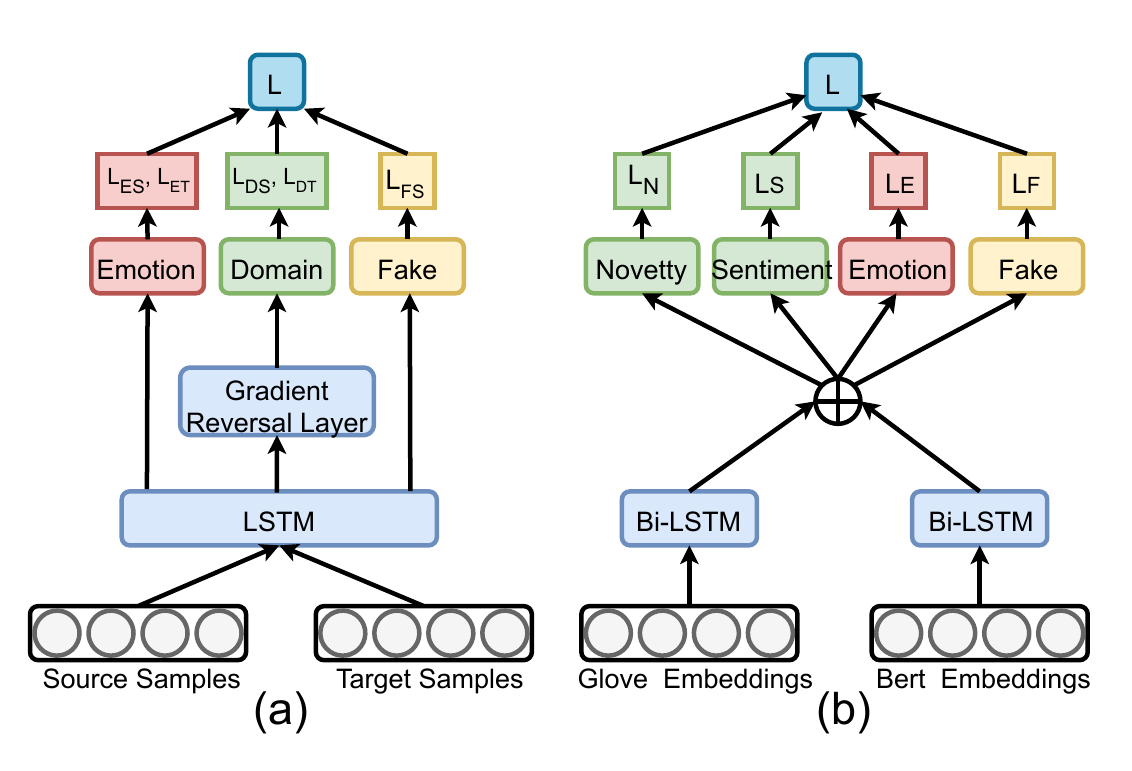}
\caption{Emotion-based misinformation detection based on multi-task learning. (a)  \cite{choudhry2022emotion,choudhry2022emotion33,chakraborty2023emotion233}, (b) \cite{kumari2021multitask3}}
\label{Multitask}
\end{figure}

The method developed by Choudhry et al. \cite{choudhry2022emotion,choudhry2022emotion33,chakraborty2023emotion233} (Figure~\ref{Multitask} (a)) aims to address the issue of cross-domain robustness in determining the veracity of news articles. Generalizability of the method across different domains is achieved using a domain-adaptive framework, whose aim is to facilitate the extraction of domain-invariant features by aligning the source and target domains in the feature space. The multi-task learning setup trains an emotion classifier as an auxiliary task in parallel to a fake news detector, to try to improve the alignment between the source and target domains, while adversarial training helps to make the model robust to outliers. The emotion classifier assigns emotion labels according to Ekman's or Plutchik's emotions, with the aid of the Unison model \cite{colnerivc2018emotionunisonmodel}. 
An LSTM is used as the feature extractor, which is trained using the accumulated loss from the fake news classifier, emotion classifier and a domain classifier, the latter of which acts as a discriminator in learning domain-invariant features.     

Based on the relatedness between the tasks of detecting fake news, novelty, emotion, and sentiment, Kumari et al. \cite{kumari2021multitask3} (Figure~\ref{Multitask} (b)) developed a multi-task learning framework in which the latter three of these are treated as auxiliary tasks. Using premise-hypothesis pairs as input, the model detects whether or not the hypothesis is fake with respect to the premise.  Pre-trained and/or fine-tuned models are firstly used to determine whether the hypothesis is novel with respect to the premise, and whether or not the hypothesis and premise differ in terms of binary emotion values (i,e., sadness/joy/trust vs. anger/fear/disgust/surprise) and sentiment (positive or negative). Different Bi-LSTMs that use pre-trained GloVe and BERT-based embeddings are employed to obtain two different input textual representations, which are concatenated and used as the input to the three auxiliary tasks and the main task of fake news detection.

\subsubsection{Multimodal Methods}

\begin{figure}[!t]
\centering
\includegraphics[width=\columnwidth]{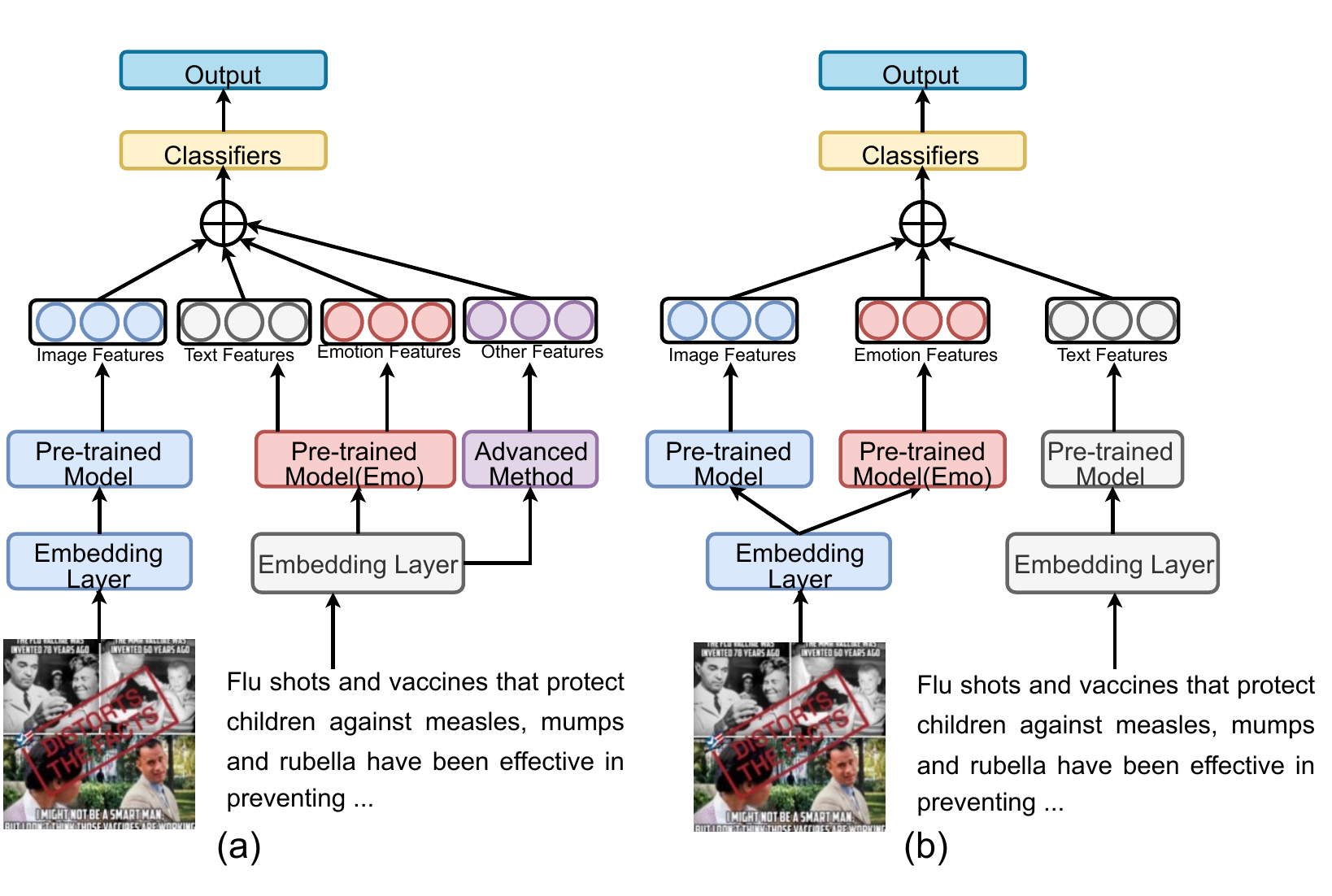}
\caption{Emotion-based misinformation detection based on temporal information. (a) Multimodel with text emotion. (b) Multimodel with image emotion}
\label{Multimodel}
\end{figure}

On platforms like Twitter or Weibo, people often attach images to their textual posts to better express their opinions or emotions. Several studies have thus attempted to exploit information from these images to improve the detection of rumors, mainly based on two different frameworks, both of which involve combining features from text and images, but which differ in terms of whether emotion features are extracted from text (Figure~\ref{Multimodel} (a)) \cite{cui2019sameaa,guo2023tiefake29} or images (Figure~\ref{Multimodel} (b))\cite{kumari2023identifying999,gupta2022mmm1111,uppada2022image5555}.

The Title-Text Similarity and Emotion-Aware Fake News Detection method \cite{guo2023tiefake29} applies BERT with a fully connected layer and ResNet-50 to obtain textual and visual features, respectively. The publisher emotion extractor from \cite{zhang2021miningwww} (described above in Section \ref{sec:dualemotion}) is used to obtain a range of emotion-based feature values from textual news content. The scaled dot-product attention mechanism is also used to capture the similarity between the title and textual features, based on the observation that authors of fake news may attempt to catch the reader’s attention by using titles that are not relevant to the news content. All features are subsequently combined and fed into a FC layer with softmax to make predictions.    

The SAME multi-modal embedding model \cite{cui2019sameaa} incorporates user sentiment for fake news detection. Firstly, VGGNet and CNN are used to represent images, while text is represented using Glove and MLP, and profiles (i.e., source, publisher and keywords) are represented using one-hot encoding. An adversarial learning mechanism is then applied to find semantic correlations between different modalities. A novel hybrid similarity loss method based on Graph Affinity Metric and Local Similarity Metric is used to incorporate the user’s sentiments (i.e., positive, negative or neutral), which are obtained using VADER. Finally, a fully connected layer with softmax is applied as a classifier.

The multimodal framework in \cite{kumari2023identifying999,gupta2022mmm1111} makes use of text and images from source-target pairs, in which the target corresponds to information from fake news datasets, while the source corresponds to background information associated with a target data item, extracted from credible websites. BERT and ResNET18 \cite{he2016deep10} are firstly used to encode the text and images of source-target pairs, respectively. The textual and visual features are concatenated to obtain multimodal feature representations. These are encoded using  VisualBERT \cite{li2019visualbert9}, which is designed to capture the rich semantics found in images and their associated text. A novelty detection module then uses these multimodal representations to determine the credibility of the new news (target) with respect to prior verified news (source), using supervised contrastive learning (SCL) such that target representations attract source representations that provide support, and repel them otherwise.  The second module pre-trains a neural network to predict image emotion labels using two classes, i.e. \textit {joy/love/sadness} vs. \textit{fear/surprise/anger}. All features are then fused and passed to MLP with softmax to make predictions.

Uppada et al. \cite{uppada2022image5555} developed a framework for fake news detection that combines visual and textual features. The architecture, which consists of two fine-tuned Xception models, makes use of the Error Level Analysis (ELA) technique to help to identify digitally altered images. One fine-tuned Xception model is trained on an ELA image dataset to detect editing traces in digital images, while the other is trained on a visual sentiment analysis dataset to determine whether images convey positive or negative sentiments. BERT is applied to learn contextual knowledge from image captions. The output of the three branches is combined and passed to the fake image classifier.

\subsection{Emotion-based stance detection in misinformation \label{sec:emo-stancedetec}}

\begin{table*}
\footnotesize
\caption{Emotion-based stance detection in rumor and fake news. EF: Emotion Features, S: Sentiment, E: Emotion, ED: Emotion Detection, ESD: Emotion-based Stance Detection, RDES: Rumor Detection based on Emotion and Stance.}
\label{tab:stance} 
\begin{tabular}{p{0.2cm}p{0.3cm}p{1.8cm}p{0.5cm}p{2cm}p{2.5cm}p{2cm}p{5cm}}

\toprule
Pub                            & Year & Data                                                           & EF     & ED                                                 & ESD                                        & RDES                        & Other Features                                                                            \\
\midrule

\cite{zeng2016unconfirmed}            & 2016 & Sydney Siege data                                              & S    & MetaMind API                                       & LR, NB, RF                                 & \-                          & Stylistic, Twitter metadata, Linguistic-informed                                          \\
\cite{wang2017ecnu55}                 & 2017 & SemEval2017                                                    & S, E & SSWE, ZSWE                                         & Ensemble Classifier                        & Ensemble Classifier         & Linguistic-informed, Stylistic, Tweet metadata, User-based,   Semantic, Cluster features, \\
\cite{bahuleyan2017uwaterloo66}       & 2017 & SemEval2017                                                    & S    & VADER                                              & XGBoost                                    & \-                          & Stylistic, Similarity, Twitter metadata, Grammatical features                             \\
\cite{akeretal2017simple100}          & 2017 & RumourEval, PHEME                                              & S    & Stanford sentiment tool                            & DT, RF, KNN                                & \-                          & Linguistic-informed, Twitter metadata, User-based, Similarity,   Lexical features         \\
\cite{enayet2017niletmrg44}           & 2018 & SemEval2017                                                    & S    & NLTK                                               & Linear SVC, LR, RF, DT, SVM                & Linear SVC, LR, RF, DT, SVM & Stylistic, Lexical, Conversation-based, User-based features                                \\
\cite{srivastava2017dfki77}           & 2018 & SemEval2017                                                    & S, E & \-                                                 & Ensemble Classifier                        & Ensemble Classifier         & Stylistic, Twitter metadata                                                               \\
\cite{hanselowski2018retrospective88} & 2018 & FNC dataset                                                    & S    & NRC-Lex, NRC-Canada                                & stacked LSTMs                              & \-                          & Linguistic-informed, Topics, Similarity features                                          \\
\cite{masood2018fake333}              & 2018 & Emergent dataset                                               & S    & Stanford Sentiment                                 & LR, RF                                     & \-                          & Grammatical, Stylistic, Structural features                                               \\
\cite{bhatt2018combining444}          & 2018 & FNC dataset                                                    & S    & lexicon based                                      & CNN, LSTM, GRU                             & \-                          & Stylistic, Linguistic-informed features                                                   \\
\cite{ghanem2019upv11}                & 2019 & SemEval2019                                                    & S, E & Various Dictionaries                               & LR                                         & LR                          & Lexical, Syntactic, Stylistic, Twitter, Conversation-based,   Cluster features             \\
\cite{janchevski2019andrejjan22}      & 2019 & SemEval2019                                                    & S    & NLTK, VADER                                        & Ensemble Classifier                        & \-                          & Stylistic, Linguistic-informed, Grammatical, Semantic,   Similarity, User-based features   \\
\cite{hamidian2019gwu33}              & 2019 & SemEval2019                                                    & S    & \-                                                 & Bi-LSTM  and rules                         & Bi-LSTM and rules           & Stylistic, Conversation-based, User-based features                                         \\
\cite{lillie2019joint99}              & 2019 & DAST                                                           & S    & Afinn                                              & LSTM, LR, SVM                              & HMMs                        & Stylistic, Lexical, Reddit metadata, Linguistic-informed,   Semantic, Similarity features \\
\cite{xuan2019rumor111}               & 2019 & SemEval2017                                                    & S    & SenticNet5                                         & LR, DT, RF, LinearSVC, NB                  & \-                          & Stylistic, Topic, User-based features                                                     \\
\cite{pamungkas2019stance555}         & 2019 & SemEval2017                                                    & E    & Various Dictionaries                               & NB, DT, SVM, RF                            & \-                          & Stylistic, Conversation-based                                      \\
\cite{giasemidis2020semi666}          & 2020 & \cite{giasemidis2016determining}; London Riots Dataset; PHEME  & S    & VADER                                              & Graph-based  Algorithm                     & \-                          & Cluster, Linguistic-informed, Lexical features                                            \\
\cite{khandelwal2021fine222}          & 2021 & SemEval2019                                                    & S, E & Various Dictionaries                               & Multi-Task Learning    based on longformer & Sentence Encoder            & Conversion-based, Stylistic, Grammatical features                                         \\
\cite{parimi2023flacorm11}            & 2023 & 2020 US Presidential Election \cite{kawintiranon2021knowledge} & S    & Topic-based Bi-LSTM \cite{baziotis2017datastories} & Fuzzy Logic                                & \-                          & Semantic, User-based features                          \\
\bottomrule
\end{tabular}
\end{table*}

In addition to emotions and sentiment, the stance of readers is also an important factor in affecting rumor diffusion. If somebody supports a piece of fake news, he/she is more likely to reshare it. Emotions can impact upon a person's thinking, judgment, and decision-making, which in turn can influence their stance toward a particular topic.  This section introduces methods that use emotion as a feature for stance detection in misinformation.

Most work in this area has been driven by shared tasks, in which a number of teams compete with each other to produce the best results for a given task and dataset.  Examples of relevant tasks include SemEval-2017-Task8 \cite{derczynski2017semeval}, SemEval-2019-Task7 \cite{derczynski2017semeval}, and the FNC dataset \cite{ferreira2016emergent-FNC,fnc12017}. Lillie et al. \cite{lillie2019joint99} constructed a Danish stance-annotated dataset (DAST), consisting of Reddit posts. A number of other publicly available stance-annotated datasets has also been used in various studies \cite{parimi2023flacorm11,zeng2016unconfirmed,giasemidis2020semi666}. Further details about these datasets are provided in Table~\ref{tab:datasum}. Most stance detection methods detect emotion features using simple dictionaries or tools, and use conventional machine learning approaches, based on sentiment features and a variety of other features. Further details are provided in Table~\ref{tab:stance}.

\section{Discussion \label{sec:disscussion}}

\begin{table*}[htb]
\footnotesize
\caption{Performance of advanced fusion methods. \textit{Ablation: No emotion}, use the same methods as the \textit{Evaluation} column, but without sentiment/emotion features. The evaluation scores provided for each dataset listed in the \textit{Data} column, separated by commas (Example: score${_{Data1}}$, score${_{Data2}}$). When separate results are provided for different categories in the dataset, these categories are shown in brackets in the \textit{Data} column. The same structure is used to report the specific results for the different categories in the scores for different categories are indicated like (score${_{True}}$, score${_{False}}$), with the categories already labeled in the Data.}

\label{tab:performance}
\begin{tabular}{p{1.3cm}|p{3cm}p{4.3cm}p{4cm}p{2.7cm}}

\toprule
Methods                                                                                              & Pub and baseline                                              & Data                                                       & Evaluation                                            & Ablation: No emotion                                      \\ 
\midrule
\multirow{7}{*}{\shortstack{Methods \\ Combining \\ Emotion \\with Other   \\Text-Based \\Features}} & FakeFlow \cite{ghanem2021fakeflow2222}                        & MultiSourceFake, LUN                                       & F1-macro 0.96, 0.96                                   & MultiSourceFake 0.91                                       \\
                                                                                                     & Baseline (BERT)                                               & MultiSourceFake                                            & F1-macro 0.93                                         & \-                                                         \\ \cline{2-5} 
                                                                                                     & MixtureofExperts\cite{zhao2023collaborative8888}         & Weibo21                                                    & F1 0.9223                                             & 0.9185                                                     \\
                                                                                                     & Baseline (BERT)                                               & \-                                                         & F1 0.8795                                             & \-                                                         \\ \cline{2-5} 
                                                                                                     & EmoAttentionBERT\cite{kelk2022automatic444}               & MultiFC (snopes, politifact)                               & F1-macro 0.344, 0.318                                 & \-                                                         \\
                                                                                                     & Baseline (BERT)                                               & \-                                                         & F1-macro 0.295, 0.282                                 & \-                                                         \\ \cline{2-5} 
                                                                                                     & LSTM with Fuzzy Sentiment\cite{mohamed2022applying777}     & Combination of Infodemic and    CoAID                      & F1 0.9143                                             & 0.9024                                                     \\ \hline
\multirow{5}{*}{\shortstack{Mining of \\Dual \\Emotions}}                                            & AGWu-RF \cite{luvembe2023dual14}                              & RumourEval19, PHEME, Fakeddit                              & F1 0.95, 0.97, 0.97                                   & \-                                                         \\
                                                                                                     & Baseline(RumorEval,PHEME:DTCA\cite{wu2020dtca1}; Fakeddit:DeepNet\cite{kaliyar2020deepnet2})& & F1 0.82, 0.83, 0.83                                   & \-                                                         \\ \cline{2-5} 
                                                                                                     & MDE   \cite{zhang2021miningwww}                               & RumourEval19, Weibo16, Weibo20                             & F1-macro 0.346, 0.867, 0.915                          & \-                                                         \\ \cline{2-5} 
                                                                                                     & EFN   \cite{guo2019exploiting33}                              & Customized                                                 & F1 0.874                                              & 0.859                                                      \\
                                                                                                     & Baseline (GRU)                                                & \-                                                         & F1 0.84                                               & \-                                                         \\ \hline

\multirow{12}{*}{\shortstack{Methods \\Based \\on Tree\\ or Graph \\Structures}}                     & SSE-BERT \cite{miao2021syntax21}                              & Weibo16, Weibo20                                           & F1 0.947, 0.943                                       & 0.941, 0.94                                                \\
                                                                                                     & Baseline (Bi-GCN)                                             & \-                                                         & F1 0.892,  0.882                                      & \-                                                         \\ \cline{2-5} 
                                                                                                     & ptVAE   \cite{fang2023unsupervised11}                         & Weibo16,Twitter15, Twitter16(True,False)                & F1 (0.853,0.848), (0.67,0.697), (0.682,0.682)         & (0.776,0.754),   (0.6,0.638),(0.641,0.676)                \\
                                                                                                     & Baseline (GFVAE   \cite{ma2021gf4})                           & \-                                                         & F1(0.752,0.745),(0.623,0.653),(0.639,0.648)&                                               \\ \cline{2-5} 
                                                                                                     & \cite{zhang2023sentiment33333}                                & Weibo16                                                    & F1 0.97                                               & \-                                                         \\
                                                                                                     & Baseline (BERT)                                               & \-                                                         & F1 0.88                                               & \-                                                         \\ \cline{2-5} 
                                                                                                     & MHN   \cite{zhang2023sentence333}                             & LUN, SLN                                                   & F1-macro 0.7169, 0.8972                               & (LUN) no sentiment net 0.6983 \\
                                                                                                     & Baseline (GCN+Attn)                                           & \-                                                         & F1-macro 0.6642, 0.8524                               & \-                                                         \\ \cline{2-5} 
                                                                                                     & SA-HyperGAT   \cite{dong2022sentiment1hh}                     & Twitter15, Twitter16 (UR, NR, TR, FR)                      & F1 (0.857,0.838,0.923,0.88), (0.925,0.886,0.957,0.86) & (0.837,0.763,0.905,0.861),   (0.866,0.765,0.939,0.87)      \\
                                                                                                     & Baseline (Bi-GCN)                                             & \-                                                         & F1   (0.752,0.772,0.885,0.847), (0.818,0.772,0.885,0.847)&                                      \\ \cline{2-5} 
                                                                                                     & GCS   \cite{haque2022graph3333}                               & PHEME                                                      & F1 0.9342                                             & \-                                                         \\
                                                                                                     & Baseline   (\cite{ajao2019sentiment23})                       & \-                                                         & F1 0.8496                                             & \-                                                         \\ \hline
\multirow{6}{*}{\shortstack{Methods \\Based on \\Temporal \\Information}}                            & TDEI \cite{wang2021rumor151}                                  & Weibo16 (True, False)                                      & F1 (0.969,0.968)                                      & (0.959,0.958)                                              \\
                                                                                                     & Baseline(RVNN)                                                & \-                                                         & F1 (0.911,0.905)                                      & \-                                                         \\ \cline{2-5} 
                                                                                                     & SD-TsDTS-CGRU   \cite{wang2019rumor2789123,wang2020rumor}     & Weibo16, Twitter16-2 (non-rumor,rumor)                   & F1 (0.963,0.963), (0.880,0.889)                       & Weibo16(rumor) 0.92                                        \\
                                                                                                     & Baseline(GRU)                                                 & \-                                                         & F1 (0.830,0.835), (0.796,0.804)                       & \-                                                         \\ \cline{2-5} 
                                                                                                     & RvNNwithTemporal\cite{fu2022rumor7777}                    & Weibo16, RumourEval19                                      & F1-macro 0.939, 0.534                                 & 0.925,0.492                                                \\
                                                                                                     & Baseline(RVNN)                                                & \-                                                         & F1-macro 0.919, 0.506                                 & \-                                                         \\ \hline
\multirow{4}{*}{\shortstack{Multitask \\Learning}}                                                   & \cite{choudhry2022emotion}                                    & PHEME, FakeNewsAMT, Celeb, Gossipcop                      & F1 0.864, 0.866, 0.879, 0.778                         & 0.848, 0.806, 0.815,0.745                                  \\ \cline{2-5} 
                                                                                                     & \cite{choudhry2022emotion33,chakraborty2023emotion233}        & Source: FakeNewsAMT; Target:   Gossipcop (cross domain)    & Accuracy 0.795                                        & 0.451                                                      \\ \cline{2-5} 
                                                                                                     & \cite{kumari2021multitask3}                                   & ByteDance, FNC, Covid-Stance                               & F1 0.9974, 0.9688, 0.9859                             & 0.8821, 6826, 0.8428                                       \\
                                                                                                     & Baseline(SiameseLSTM\cite{shih2017investigating3})         & \-                                                         & F1 0.8783, 0.675, 0.8392                              & \-                                                         \\ \hline
\multirow{10}{*}{\shortstack{Multimodal \\Methods}}                                                  & \cite{guo2023tiefake29}                                       & PolitiFact, GossipCop                                      & F1 0.92, 0.894                                        & 0.914, 0.892                                               \\
                                                                                                     & Baseline (BERT)                                               & \-                                                         & F1 0.818, 0.850                                       & \-                                                         \\ \cline{2-5} 
                                                                                                     & \cite{cui2019sameaa}                                          & PolitiFact, GossipCop                                      & F1-macro 0.7724, 0.8042                               & 0.7085, 0.7091                                             \\
                                                                                                     & Baseline (SVM)                                                & \-                                                         & F1-macro 0.6557, 0.6124                               & \-                                                         \\ \cline{2-5} 
                                                                                                     & \cite{gupta2022mmm1111}                                       & MMM (real, fake)                                           & F1 (0.960,0.949)                                      & 0.926, 0.907                                               \\
                                                                                                     & Baseline(MLBERT+ResNet)                                      & \-                                                         & F1 (0.765,0.703)                                      & \-                                                         \\ \cline{2-5} 
                                                                                                     & \cite{kumari2023identifying999}                               & NovEmoFake                                                 & F1-micro 0.9775                                       & 0.9054                                                     \\
                                                                                                     & Baseline(BERT,ResNet)                                       & \-                                                         & F1-micro (0.8002, 0.7401)                             & \-                                                         \\ \cline{2-5} 
                                                                                                     & \cite{uppada2022image5555}                                    & Fakeddit                                                   & F1 0.9329 Accrucy 0.9194                              & \-                                                         \\
                                                                                                     & Baseline(BERT+ResNet)                                        & \-                                                         & Accrucy 0.8909                                        & \-                                                         \\ 
                                                                                                     
\bottomrule
\end{tabular}
\end{table*}

The analysis in Section \ref{sec:emotion-basedrumordetec} revealed the wide range of designs of advanced fusion methods for emotion-based misinformation detection. In this section, we conduct a comparative analysis to identify the most effective strategies. Table \ref{tab:performance} provides performance statistics in terms of F1-score for a range of the advanced methods discussed in Section \ref{sec:emotion-basedrumordetec}, along with those of the baseline methods used for comparison. Where possible, we also provide the results of ablation experiments,  i.e., where sentiment/emotion features are excluded to assess their impact on overall performance.  As part of our analysis, we compare the performance of different methods that have been evaluated using the same dataset. Although we suggest possible reasons for different performance levels, it is important to note that performance may influenced by numerous factors, including differences in data processing methods, selection of base models, and the predictive behavior of the sentiment model, etc. 

\textbf{Feature Fusion:}  Textual data contains an abundance of information, which has been encoded using a wide variety of features in different misinformation detection methods, as illustrated in Figure~\ref{detectionfeatures}. In Table~\ref{tab:methods}, we list the specific features that have been combined with emotion and/or sentiment information in different studies. While a comparison of the performance of complete models with those of ablation experiments in Table \ref{tab:performance} confirms the importance of sentiment and emotion features in misinformation detection, high levels of performance can only be achieved by combining multiple features. For example, it is shown in  \cite{ghanem2021fakeflow2222} that the proposed combination of topic and affective features outperforms the use of either topic or affective features in isolation. Furthermore, \cite{kumari2023identifying999,gupta2022mmm1111} show that extracting features from the images that accompany posts on social media platforms can provide additional clues about the emotional states and behaviors of individuals and thus help to boost the results of misinformation detection.

\textbf{Model Fusion:} Different models and learning techniques have their own advantages and disadvantages, and optimal misinformation detection performance methods can generally only be achieved by combining a number of different techniques. For example, pre-trained models like GloVe and  BERT  are effective in encoding textual content with word embeddings, while RoBERTa can be successfully employed for sentiment and emotion detection \cite{mao2022biases2}. Meanwhile, methods like CNN, LSTM, or GRU may be usefully adopted for feature extraction. Encoding information about the graph structure inherent in many datasets requires different approaches. For instance, dependency and sentiment trees may be used to represent grammatical or semantic aspects of sentence structure, while GCN and hypergraphs can encode the tree-like structure of social media data. To capture temporal features of rumor propagation, different studies have utilized RNN, LSTM and GRU models, which excel in handling time series data. The application of fusion or ensemble techniques can fully leverage the relative strengths of these different types of methods and models, as may be confirmed by comparing the results of the advanced methods with baseline methods in Table \ref{tab:performance}.

\textbf{Comparison between different fusion methods:} Although the experimental results for EmoAttention BERT \cite{kelk2022automatic444} highlight the importance of emotionally charged style, and in particular emotional intensity, as a predictive feature of fake news, Table \ref{tab:performance} illustrates that the performance of this method is low. A probable reason is that their method is evaluated on a complex dataset that includes multiple domains and labels, but their fairly simple framework fails to account for potential differences in the characteristics of data across different domains. A possible solution is to adopt a multi-task architecture, similar to \cite{choudhry2022emotion,choudhry2022emotion33,chakraborty2023emotion233}, in which a domain classifier is incorporated as a discriminator to ensure that the model performs well across multiple domains through reinforcement learning. Similarly, The challenging characteristics of the RumourEval-2019 dataset (i.e., low inter-annotator agreement and sparse data \cite{gorrell2019semeval,li2019eventai}) resulted in relatively low performance from RvNN with Temporal \cite{fu2022rumor7777} (0.53 F1) and MDE \cite{zhang2021miningwww}  (0.35 F1) methods, even though they achieved much higher results on other datasets. In comparison, AGWu-RF \cite{luvembe2023dual14} attained vastly superior results on RumourEval-2019 (0.95 F1). In common with MDE, AGWu-RF uses dual emotion features. However, its combination of these features with a random forest with genetically adapted weights appears to make it robust in handling this problematic dataset. Furthermore, AGWu-RF is demonstrated to be sufficiently generalizable for successful application to social media datasets with varying characteristics, e.g., it outperforms both \cite{choudhry2022emotion} and \cite{haque2022graph3333} on the PHEME dataset. It is also notable that while AGWu-RF uses only textual information, it achieves better results than \cite{uppada2022image5555} on the multimodal Fakeddit dataset, even though the latter method uses both text and image features.

Source tweets in Twitter15 and Twitter16 are annotated with four class labels, i.e., non-rumor (NR), false rumor (FR), true rumor (TR), and unverified rumor (UR). While these datasets are used to evaluate both the ptVAE \cite{fang2023unsupervised11} and SA-HyperGAT \cite{dong2022sentiment1hh} methods, the evaluation of ptVAE uses only two classes, i.e., \textit{true} (non-rumors and true rumors) and \textit{false} (false rumors).  Nevertheless, ptVAE exhibits lower performance than SA-HyperGAT on these datasets, and is also inferior to several other methods that have been evaluated on the Weibo16 dataset.  This could be due to the less rigorous data processing methods used in ptVAE, but it is more likely that their proposed VAE architecture for sentiment analysis is not as effective as other methods, such as the fine-tuned RoBERTa model used in SA-HyperGAT \cite{dong2022sentiment1hh}. Table \ref{tab:performance} shows that many methods use Weibo16 for evaluation. The performance comparison provides strong evidence that combining sentiment/emotion features with those accounting for the propagational and/or temporal features is highly important. Specifically, \cite{zhang2023sentiment33333}, TDEI \cite{wang2021rumor151} and SD-TsDTS-CGRU \cite{wang2019rumor2789123,wang2020rumor} and RvNN with Temporal \cite{fu2022rumor7777} all achieve high levels of performance on Weibo16 (0.94 F1 or higher). The impressive F1 of 0.95 achieved by SSE-BERT \cite{miao2021syntax21} on the same dataset, by combining sentiment and dependency tree information from source posts indicates the potential value of considering syntactic information. While MDE \cite{zhang2021miningwww} and ptVAE \cite{fang2023unsupervised11} perform the worst, with scores below 0.9. Additionally, both FakeFlow \cite{ghanem2021fakeflow2222} and MHN \cite{zhang2023sentence333} were evaluated on the LUN dataset, and analyze changes in affective information across the different parts of articles. However, the superior performance of FakeFlow suggests that accounting for affective interactions with different topics represents a more successful approach.

The analysis above underlines the complexities of developing effective misinformation detection models. High levels of performance can only be achieved through leveraging multiple relevant features, which include thematic, temporal, propagation structure, dual emotion and/or image information, in addition to sentiment and emotion. Furthermore, the specific methods chosen to learn or represent these features can also impact upon performance, and it is usually necessary to combine a range of learning methods to achieve optimal results. Moreover, to achieve cross-domain robustness, the employment of multi-task learning frameworks incorporating reinforcement learning can be advantageous.  

\section{Challenges and future research directions \label{sec:challengeanddirection}}

While this article has reviewed a large and diverse body of research relating to emotion-based misinformation detection, there still remains a variety of unsolved challenges in this field.  In this section, we outline the most important of these challenges, and discuss potential future directions of research.

\subsection{Dataset Collection (Multi-platform, Multilingual)}

There are many popular social media platforms such as Twitter, Facebook, Reddit and Sina Weibo, among others, which constitute major means of spreading misinformation. While the language used on each platform is diverse, and the data formats are varied, making the processing of such data cumbersome. Given that the dissemination of fake news is a global problem, it is important to develop approaches that are more universally applicable than most currently available methods. However, achieving this goal is hindered by the limitations of the majority of currently publicly available datasets, which are usually collected from a single platform (as shown in Table~\ref{tab:datasum}) and which predominantly concern textual data in a single language (typically English or Chinese). Only by developing larger and more diverse datasets will it be possible to develop more general models that are urgently needed. These should cover multiple data formats obtained from different platforms and covering multiple languages.

\subsection{Annotation (Emotion)}

The development of emotion-based misinformation detection methods with optimal performance requires that supporting misinformation datasets are annotated with reliable emotion and/or sentiment labels, since inaccurate labels are likely to impact negatively on the overall performance of the methods. While this is most often carried out using dictionary lookup, some studies have employed transfer learning methods, by applying models trained on other sentiment analysis or emotion-labeled datasets to automatically annotate the emotions expressed in misinformation datasets. Examples include \cite{choudhry2022emotion,choudhry2022emotion33,chakraborty2023emotion233}, which use a previously developed Unison model, and \cite{dong2022sentiment1hh}, which utilizes a fine-tuned RoBERTa model. However, the
emotion labels obtained in these ways are not sufficiently accurate. Compared to time-consuming manual annotation, a more promising approach is to use LLMs to annotate emotion and/or sentiment \cite{zhang2023enhancing,feng2023affect2,lei2023instructerc1}, given their advanced capabilities and transferability. Both  \cite{feng2023affect2} and \cite{lei2023instructerc1} have demonstrated that LLMs can compete with or exceed the state-of-the-art (SOTA) in recognising emotions in dialogue. In particular \cite{feng2023affect2} showed that the LLaMA-7B \cite{touvron2023llama5} model can achieve performance levels close to those of SOTA supervised methods, but using only half as much training data for fine-tuning. 
Zhang et al. \cite{zhang2023enhancing} developed an instruction-tuned LLM for financial sentiment analysis which, augmented with additional context from external sources, is able to outperform LLM baselines such as ChatGPT and LLaMA by margins of between 15\% and 48\%. The above studies all highlight the tremendous potential of LLMs in the field of sentiment analysis.

\subsection{Multimodality}

Although rumors and fake news were traditionally spread through face-to-face communication, the emergence of social media resulted in their primary means of dissemination switching to text. However, continual advances in technology have led to an increasing shift towards multimodality. For example, people now frequently augment textual post content with images or videos,  while on platforms like YouTube or TikTok, videos are the predominant means of sharing information. Accordingly, it is becoming increasingly important to explore methods that can address the challenges of multimodality\cite{comito2023multimodal8}, and that are able to adapt to the ever-changing characteristics of social media communication. While we have reviewed a number of approaches that combine text and image-based information, recent advanced multi-modal models that integrate language and visual understanding provide considerable scope for further research in this area. For example, GPT-4 has a certain level of visual understanding capability, although its implementation details have not been publicly disclosed. Inspired by the success of LLMs, some studies have started to focus on large multi-modal models, such as LLaVA \cite{liu2023visual1,liu2023improved2}, an end-to-end large multimodal model that connects a visual encoder and a large language model to achieve general visual and language understanding.  Additionally, MiniGPT-5 \cite{zheng2023minigpt3} introduces a novel interleaved vision-and-language generation technique, with a focus on non-descriptive multi-modal generation. Exploring the integration of these large multimodal models within misinformation detection methods is an interesting and promising research direction.

\subsection{Benchmark}
Several benchmark datasets have been developed in the context of shared tasks, which are aimed at evaluating various different characteristics of misinformation detection methods. For example, the datasets created for SemEval-2017-Task8 \cite{derczynski2017semeval} and SemEval-2019-Task7 \cite{derczynski2017semeval} focus on stance detection and misinformation detection, while the CLEF2020 - CheckThat! Lab \cite{clef2020-checktahtlab} addresses multilingualism through the inclusion of benchmark data in both English and Arabic. These are complemented by a recently developed multi-modal benchmark for fake news detection \cite{qi2023fakesv1}. Despite the value of these datasets in facilitating the evaluation of various different \textit{individual} aspects of methods, there is still a lack of a suitably comprehensive benchmark that can \textit{simultaneously} evaluate the ability of misinformation detection methods to handle diverse types of multi-modal data from multiple platforms and covering different languages, as well as assessing their ability to perform important subsidiary tasks such as emotion, sentiment and stance detection, identification of rumor source, etc. We believe that the development of such a dataset would be of enormous value in helping to guide research in this area towards the development of more robust and universally applicable misinformation methods, as well as focusing attention on the development of important supporting technologies.

\subsection{Interpretability}

Understanding how and why misinformation detection models have arrived at their decision about whether or not a post or news article represents true or fake information can be important to make their reasoning processes more transparent and make it easier to understand why errors occur. However, despite the high levels of performance achieved by many DL approaches, their black-box nature means that no such reasoning information is available, and that their decisions are hard to justify. Although it remains a challenge to develop models that are both sufficiently accurate \textit{and} whose results are interpretable, several studies have proposed possible solutions for explainable misinformation detection. These include the use of topic-based features for classification \cite{hosseini2023interpretable1}, Explainable Artificial Intelligence (XAI) techniques \cite{dua2023flash2} and Commonsense Knowledge Graphs \cite{gao2023interpretable3}. Recent research has also begun to focus on the development of interpretable LLMs \cite{zhao2023explainability5}, such as MentalLLaMA \cite{yang2023mentalllama4}, which is an interpretable mental health analysis model based on LLaMA-2. Accordingly, it is hoped that researchers working in misinformation detection will begin to place greater emphasis on exploring the increasing range of options that could be used to improve the interpretability of their models.    

\subsection{Large Language Models}

The popularity of ChatGPT and GPT-4 \cite{openai2023gpt4} has resulted in the powerful capabilities of LLMs becoming widely known \cite{zhao2023survey7}. As mentioned above, there is potential for LLMs to be employed in misinformation detection in multiple ways, including sentiment and emotion detection, multimodal analysis, and to enhance the interpretability of detection models. Some studies have additionally begun to explore the use of LLMs for rumor and fake news prediction. For example, Hu et al. \cite{hu2023bad1} designed a framework for fake news detection in which a small language model (i.e., BERT) is complemented by an LLM, which provides multi-perspective guiding principles to improve prediction accuracy. Meanwhile, Pavlyshenko et al. \cite{pavlyshenko2023analysis2} designed prompts to fine-tune LLaMA for rumor and fake news detection. Cheung et al \cite{cheung2023factllama3} used external knowledge to bridge the gap between knowledge encoded in the LLM and the most up-to-date information available on the Internet, in order to enhance fake news detection performance. The promising results achieved by these approaches, combined with the indisputable power and advanced capabilities of LLMs, motivate further exploration of how they can be best exploited to further improve the accuracy of rumor and fake news detection.

\section{Conclusion \label{sec:conclusion}}

The unstoppable growth of social media is making it easier than ever for misinformation to spread rapidly and widely.  As such, there is an increasingly urgent need for robust automated methods that can detect and stop this spread as efficiently and effectively as possible. In this article, we have comprehensively analyzed emotion-based applications for rumor and fake news detection. After introducing related work, we firstly motivated such approaches by summarizing research that confirms the strong links between emotion and misinformation. We subsequently provided an overview of available datasets that can support the development of misinformation detection methods, followed by a summary of both conventional and deep learning methods that have been employed in emotion-based misinformation detection approaches. We then proceeded to describe and categorise a diverse range of recently proposed advanced methods that combine the use of emotion and/or sentiment with various other features, and which integrate a number of different learning methods to achieve their goals. We additionally provided an overview of emotion-based stance detection methods in misinformation. Subsequently, we discussed the relative strengths and weaknesses of different advanced methods from various perspectives. Finally, we outlined several unsolved challenges in the field of rumor and fake news detection, and provided suggestions for future research directions, with a focus on the greater exploitation of the increasingly ubiquitous LLMs.  In summary, our review has aimed to demonstrate the significant role of sentiment and emotion in misinformation, and to highlight the most important aspects in its automated detection.  It is intended that the survey will enable researchers who are interested in this field to better appreciate the potential value of affective information in misinformation detection, and will help to drive further advances to the SOTA in this field.


\appendix


\section{Specific types of content-based features \label{appendix:features}}

\begin{table}
\footnotesize
\caption{Specific types of content-based features}
\label{tab:contentbasedfeatures}
\begin{tabular}{p{1.4cm}p{6.5cm}}
\toprule
Term                         & Features                                                                                                                                                                                                                                                                                                                                                                                                                 \\
\midrule
Similarity Features          & title-text   similarity, word similarity, sentence similarity, cosine similarity between source post and related comments                                                                                                                                                                                                                                                                                                              \\
Cluster Features             & word-cluster feature,  brown cluster feature \cite{akeretal2017simple100}, SDQC depth-based clusters \cite{akeretal2017simple100}                                                                                                                                                                                                                                                                                                                                                 \\

Semantic Feature             & word vector features   (Glove \cite{pennington2014glove}, BERT \cite{devlin2018bert}, GoogleW2V \cite{mikolov2013distributed}, Word2vec \cite{mikolov2013efficient})                                                                                                                                                                                                                                                                                                                                                                \\

Grammatical Features         & part-of-speech tags, noun, verbs, adjectives, and pronouns                                                                                                                                                                                                                                                                                                                                                             \\

Lexical Features             & bad sexual words, cue words, multilingual hate lexicon, linguistic words, specific categories, denial term, support words, negation words, swear words, surprise and doubt words                                                                                                                                                                                                                                   \\

Linguistic-informed Features & tf-idf, n-gram, named entity recognition, text language, bag-of-characters, bag of words (BoW)                                                                                                                                                                                                                                                                                                                          \\

Stylistic Features\cite{ghanem2019upv11}           & question marks, exclamation marks, punctuation marks, length of a sentence, uppercase ratio, consecutive characters and letters, presence of URLs, number of stop words, number of upper case letters, number of lower case letters, number of numeric values, word count, character count, sentence  count, average sentence length, ease of comprehension, lexical  diversity \\

Syntactic Features           & ratio of negation, bag of relations (all tokens, list of words, verbs)                                                                                                                                                                                                                                                                                                                                                  \\

Conversation based Features  & text similarity to source tweet, text similarity to replied tweet, tweet depth                                                                                                                                                                                                                                                                                                                                         \\

Twitter Metadata \cite{al2023exploring,janchevski2019andrejjan22}             & the number of characters in a tweet, the number of retweets, favorites, presence of hashtags, URLs, mentions, existence of photos, creating time gaps for posts, Twitter verification. etc.                                                                                                                                                                                                                                                \\

Reddit Metadata \cite{lillie2019joint99}              & karma, gold status, Reddit employment status (if any), verified e-mail, reply count, upvotes, and whether the user is the submission submitter. Reddit commenting syntax:   sarcasm (‘/s’), edited (‘edit:’), and quote count (‘>’)                                                                                                                                                                                  \\

Others                       & Topics, term   features, textual novelty   \\                                                                        \bottomrule                                                               
\end{tabular}
\end{table}

\section{Emotion Detection Tools \label{appendix:EmotionsExtractedTools}}

\begin{table*}
\footnotesize
\caption{Emotion Detection Tools }
\label{EmotionsExtractedTools}
\begin{threeparttable}
\begin{tabular}{p{2.7cm}p{1.2cm}p{9.5cm}p{3cm}}
\toprule
Tool                                                                             & Target Language & Description                                                                                                                                                                       & Function                                               \\
\midrule
SenticNet \cite{cambria2022senticnet7}                                           & English      & Concept-level lexicon leveraging the denotative and connotative information associated with words and multi-word expressions.                                               & sentiment, intensity, emotion                    \\
NRC Emotion Lexicon (EmoLex) \cite{mohammad2013crowdsourcingnrc}                 & English      & Crowd-sourced lexicon associating words with emotions and sentiments.                                                                                                                                            & sentiment, emotion                                     \\
NRC Intensity Emotion Lexicon\cite{LREC18-AILIntensities}                        & English      & Lexicon associating words with real-valued intensity scores                                                                               & emotion, intensity                                     \\
AraNet \cite{abdul2019aranet}                                                    & Arabic       & Collection of BERT-based social media processing tools predicting various types of information including emotion, irony and sentiment                                                                                                                                                   & sentiment, emotion, irony                                \\
CAMeL\cite{obeid2020camel}                                                       & Arabic       & An open-source package consisting of a set of Python APIs for NLP with accompanying command-line  tools that thin-wrap these APIs                                                   & sentiment                                              \\
Affective Lexicon Ontology (ALO)\cite{xu2008constructingALO}                     & Chinese      & Lexicon in which each entry is with an emotion and sentiment polarity                                          & sentiment, emotion                                     \\
TextBlob\tnote{a}                                         & English      & A Python sentiment   analysis library that uses the Natural Language ToolKit (NLTK)                                                                                                   & sentiment scores with subject and polarity              \\
LIWC\cite{pennebaker2015developmentliwc}                                         & Multilingual & Text analysis software to conduct various calculations related to emotions,   social dynamics, and cognitive processes by counting relevant words.              & Various text analyses                                  \\
Valence Aware Dictionary for sEntiment Reasoning (VADER)   \cite{hutto2014vader} & English      & An open-source   rule-based sentiment analysis tool suitable for analyzing social media   text                                                                                    & sentiment with score                                   \\
Sentilex-PT02\tnote{b}                             & Portuguese   & A sentiment lexicon   for Portuguese, consisting of 7,014 lemmas, and 82,347 inflected forms                                                                                         & sentiment                                              \\
AFINN\cite{nielsen2011newAfinn}                                                  & English      & An open source dictionary-based sentiment analysis tool, which assigns numerical sentiment polarity scores.                                                                & sentiment with score                                   \\
cn-sentiment-measures\tnote{c}                        & Chinese      & A toolkit for estimating Chinese sentiment scores based on multiple measures.                                                                                                       & sentiment with score                                   \\
EmoSenticNet (EmoSN)\cite{poria2013enhancedEmoSenticNet}                         & English      & A enriched   version of SenticNet, consisting of 13,189 words labeled according to Ekman’s six basic emotions                                                                           & sentiment, emotion                                     \\
SentiStrength\cite{thelwall2010sentiment}                                        & English      & A sentiment strength   detection algorithm which uses a lexical approach that exploits a list of   sentiment-related terms                                                & sentiment with strength                  \\
SentiWordNet\cite{esuli2007sentiwordnet}                                         & English      & A publicly available  lexical resource that associates each WordNet synset with three numerical scores denoting objectivity, positivity, and negativity)                                               & sentiment with score                                   \\
HowNet\cite{dong2003hownet}                                                      & Bilingual    & An online   common-sense knowledge base containing English and Chinese words that identifies inter-conceptual relations and inter-attribute relations of concepts  & sentiment scores, sentimental words and degree   words \\
SentiSense\cite{de2012sentisense}                                                & English      & An lexicon that attaches emotional meanings to WordNet synsets using 14 categories.                                                                                                      & sentiment, intensity, emotion                          \\
Affective Norms for English Words (ANEW)   \cite{bradley1999affective}           & English      & Lexicon of words rated by humans according to the Valence-Arousal-Dominance (VAD) model                                                                                     & Valence, Arousal, Dominance                   \\
\bottomrule
\end{tabular}
\begin{tablenotes}
      \item[a] https://textblob.readthedocs.io/
      \item[b] https://b2find9.cloud.dkrz.de/dataset/b6bd16c2-a8ab-598f-be41-1e7aeecd60d3
      \item[c] https://github.com/dhchenx/cn-sentiment-measures
    \end{tablenotes}
\end{threeparttable}
\end{table*}

Various tools and resources are used for the detection of emotion and sentiment features, the most commonly used of which are summarized in Table~\ref{EmotionsExtractedTools}.
In addition to the methods in Table~\ref{EmotionsExtractedTools}, there are also some other efficient sentiment analysis tools such as 
Emojis Dictionary\footnote{https://drive.google.com/file/d/1G1vIkkbqPBYPKHcQ8qy0G2zkoab\\2Qv4v/view}, 
Emoticons list\footnote{https://en.wikipedia.org/wiki/List\_of\_emoticons}, 
Affect-Br\cite{carvalho2018affectpt}, 
SemEval\footnote{http://www.saifmohammad.com/WebPages/SCL.html}, 
MPQA\footnote{http://www.purl.org/net/ArabicSA}, 
ENGAR\cite{liu2005opinion}, 
Hespress Facebook\footnote{https://fr-fr.facebook.com/Hespress}, 
Offense lexicon\footnote{https://sites.google.com/site/offensevalsharedtask/}, 
Sarcasm lexicon\cite{hansen2019neural}, 
Named entities lexicon (Religion lexicon, Nationality lexicon, Named entities)\cite{touahri2020evolutionteam222}, 
Baidu sentiment API\footnote{https://ai.baidu.com/tech/nlp\_apply/sentiment\_classify},
NLTK sentiment module\footnote{https://www.nltk.org/api/nltk.sentiment.html?highlight=sentiment\\\#module-nltk.sentiment},
SEO Scout’s analysis tool\footnote{https://seoscout.com}, 
IBM Watson’s Natural Language Understanding (NLU)\footnote{https://https://www.sciencedirect.com/topics/computer-science/natural-language-understanding},  
MeaningCloud\footnote{https://www.meaningcloud.com/}, 
ParallelDots\footnote{https://apis.paralleldots.com/text\_docs/index.html },
Empath\cite{fast2016empath},
EffectWordNet\cite{choi2014effectwordnet},
Hu\&Liu opinion lexicon\footnote{http://www.cs.uic.edu/liub/FBS},
SSWE\cite{tang2014learning}, 
NRC-Canada\cite{mohammad2013nrc},
Stanford sentiment Tree\cite{socher2013recursive},
Dictionary of Affect in Language (DAL) \cite{whissell2009using},
Affective Norms for English Words (ANEW) \cite{bradley1999affective},
MetaMind sentimentclassifier API\footnote{https://www.metamind.io}.

\section{Evaluation Measurements \label{appendix:EvaluationMeasurements}}

\subsection{Misinformation Detection Evaluation}

A variety of techniques has been used to evaluate the output of misinformation detection methods, including accuracy, recall, precision, and F1-score, Macro F1, class-wise F1-score, AUC \cite{hamed2023fake46,ali2023rumour555,zhao2023collaborative8888}, and RMSE \cite{zhang2021miningwww}.
These are calculated on the basis of a number of basic concepts, which are defined as follows: TP (True Positive) refers to the number of samples that the model correctly predicts as positive; TN (True Negative) refers to the number of samples that the model correctly predicts as negative; FP (False Positive) refers to the number of samples that the model incorrectly predicts as positive; FN (False Negative) refers to the number of samples that the model incorrectly predicts as negative.

The accuracy indicates the overall classification correctness of a model:
\begin{equation}
Accuracy = \frac{TP + TN}{TP + FP + TN + FN}
\end{equation}
Recall measures the model's ability to identify positive-class samples:

\begin{equation}
Recall = \frac{TP}{TP+FN}
\end{equation}

Precision measures the proportion of true positive samples among the samples predicted as positive by the model:

\begin{equation}
Precision = \frac{TP}{TP+FP}
\end{equation}

F1 score takes into account both precision and recall and represents the harmonic mean of precision and recall:

\begin{equation}
F1_{score} = \frac{2*Recall*Precision}{Recall+Precision}
\end{equation}

Macro F1 is used to evaluate the performance of multi-class classifier, by combining the F1$_{score}$ of each class; Class-wise F1$_{score}$ refers to the F1$_{score}$ of each individual class, and can be used to evaluate the performance of the classifier for each class. The AUC (Area Under the Curve) is a commonly used metric for evaluating the performance of classification models. It measures the predictive ability of a model by calculating the area under the ROC (Receiver Operating Characteristic) curve.

Root Mean Squared Error (RMSE) represents the expected value of the squared error.
\begin{equation}
RMSE = \sqrt {\frac{1}{n}\sum\limits_{i = 1}^n {{{\left| {{y_i} - {{\hat y}_i}} \right|}^2}} }
\label{RMSE}
\end{equation}
\subsection{Stance Detection Evaluation Measurements}

Accuracy, recall, precision, and F1-score, Macro F1, class-wise F1-score, FNC1-Score \cite{masood2018fake333,bhatt2018combining444}, weighted accuracy \cite{giasemidis2020semi666} are used in stance detection.

The FNC-1 weighted accuracy score is used as the final evaluation metric for the FNC-1 dataset.

\begin{equation}
\begin{split}
FNC-1_{score} = 0.25*Accuracy_{Unrelated} + \\
0.75*Accuracy_{Agree,Disagree,Discuss}
\end{split}
\end{equation}

Weighted Accuracy is a performance metric that takes into account the weight of each class in an imbalanced dataset. It calculates the overall performance of the model by taking a weighted average of the accuracy for each category. 

\printcredits

\bibliographystyle{elsarticle-num}

\bibliography{cas-refs}




\end{document}